\documentclass[review]{elsarticle}
\usepackage[table]{xcolor}

\usepackage{lineno,hyperref}
\modulolinenumbers[5]

\usepackage{lscape}
\usepackage{adjustbox}
\usepackage{lmodern}
\usepackage{amsmath} 
\usepackage{tikz}
\usetikzlibrary{tikzmark, calc} 
\usepackage{xcolor}

\biboptions{authoryear}

\usepackage{booktabs}
\usepackage{graphicx}
\usepackage{caption}
\usepackage{subcaption}
\usepackage{xcolor}
\usepackage{multirow}
\usepackage{tikz}
\usepackage{url}
\usepackage{hyperref}

\makeatletter

\newcommand*{\getnamereftext}[1]{%
    \@ifundefined{r@#1}{}{%
    \unexpanded\expandafter\expandafter\expandafter{%
      \expandafter\expandafter\expandafter\@thirdoffive\csname r@#1\endcsname
    }%
  }%
}
\makeatother

\def\checkmark{\tikz\fill[scale=0.4](0,.35) -- (.25,0) -- (1,.7) -- (.25,.15) -- cycle;} 
\begin{document}

\begin{frontmatter}

\title{Semi-supervised classification of dental conditions in panoramic radiographs using large language model and instance segmentation: A real-world dataset evaluation}

\author[1,6]{Bernardo Silva}

\author[2]{Jefferson Fontinele}

\author[3]{Carolina Letícia Zilli Vieira}

\author[4]{João Manuel R.S. Tavares}

\author[5]{Patricia Ramos Cury}

\author[1]{Luciano Oliveira\corref{cor1}}
\cortext[cor1]{Corresponding author: 
  Tel.: +55-71-99128-0179;
}
\ead{lrebouca@ufba.br}

\address[1]{Intelligent Vision Research Lab, Institute of Computing, Federal University of Bahia, Brazil}
\address[2]{Federal University of Maranhão, Maranhão, Brazil}
\address[3]{Harvard School of Public Health, Boston, MA, United States}
\address[4]{Instituto de Ciência e Inovação em Engenharia Mecânica e Engenharia Industrial, Departamento de Engenharia Mecânica, Faculdade de Engenharia, Universidade do Porto, Porto, Portugal}
\address[5]{Dentistry Faculty, Federal University of Bahia, Bahia, Brazil}
\address[6]{Federal Institute of Bahia, Bahia, Brazil}

\begin{abstract}

Dental panoramic radiographs offer vast diagnostic opportunities, but training supervised deep learning networks for automatic analysis of those radiology images is hampered by a shortage of labeled data. Here, a different perspective on this problem is introduced. A semi-supervised learning framework is proposed to classify thirteen dental conditions on panoramic radiographs, with a particular emphasis on teeth. Large language models were explored to annotate the most common dental conditions based on dental reports. Additionally, a masked autoencoder was employed to pre-train the classification neural network, and a Vision Transformer was used to leverage the unlabeled data. The analyses were validated using two of the most extensive datasets in the literature, comprising 8,795 panoramic radiographs and 8,029 paired reports and images. Encouragingly, the results consistently met or surpassed the baseline metrics for the Matthews correlation coefficient. A comparison of the proposed solution with human practitioners, supported by statistical analysis, highlighted its effectiveness and performance limitations; based on the degree of agreement among specialists, the solution demonstrated an accuracy level comparable to that of a junior specialist.

\end{abstract}

        
\end{frontmatter}

\section{Introduction}
\label{sec1}

Imaging significantly impacts dentistry, enabling specialists to identify problems that might not be visible during a clinical examination. Imaging modalities like X-rays, computerized tomography scans, and magnetic resonance imaging provide detailed views of teeth, bones, and soft tissues \citep{white2014oral}. These tools enhance the precision of diagnoses and treatments, ensuring better patient outcomes. Among the current imaging exams, radiographs are the most common in dentistry \citep{white2014oral, langlais2016exercises}, being requested to identify various pathologies like cavities, periodontal disease, impacted teeth, and bone infections \citep{chang2020deep, yuksel2021dental} and track the progress of dental treatments. 

\begin{figure}[t]
    \centering

\includegraphics[width=\linewidth]{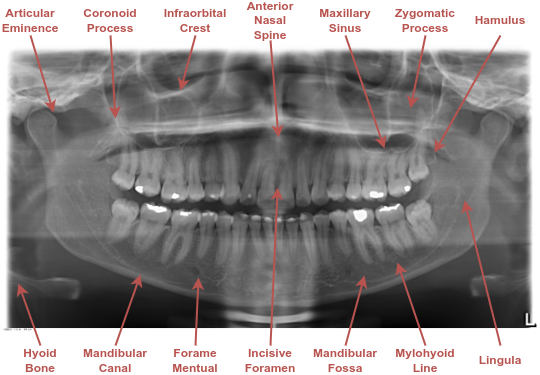}

    \caption{Sample of a human panoramic radiograph with the oral structures identified.}
    \label{fig:pan-radiographs}
\end{figure}

One of the most commonly used radiographs in dentistry is the panoramic radiograph \citep{white2014oral, langlais2016exercises, silva2018automatic}, which is an extraoral imaging technique where the X-ray film or sensor remains outside the patient's mouth during acquisition. In a single image, the panoramic radiograph provides a comprehensive view of both upper and lower jaws, but with less detail of the mouth structures \citep{haring2000dental, silva2018automatic, jader2018deep, pinheiro2021numbering}. Fig. \ref{fig:pan-radiographs} depicts an example of a panoramic radiograph, revealing the structures and their overlaps, which can lead to cluttered readings.

\begin{figure}[t]
    \centering
    \includegraphics[width=\linewidth]{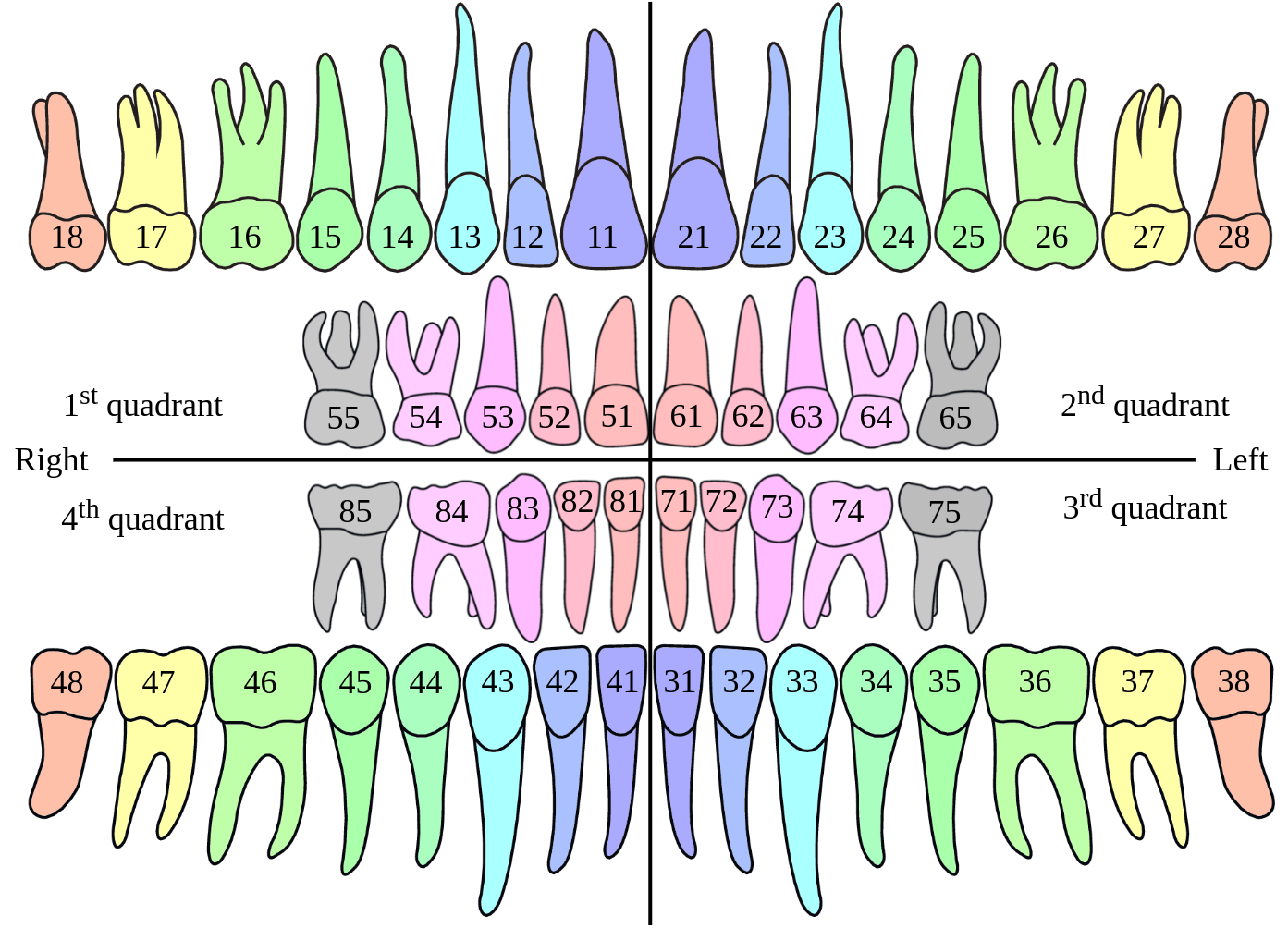}
    \caption{Illustration of FDI notation: A two-digit system that
determines each tooth (a custom color code was added to show qualitative results). Adapted from \cite{silva2023boosting}.}
    \label{fig:fdi-system}
\end{figure}

While panoramic radiography offers considerable flexibility in diagnosing periodontal bone loss, bone irregularities in the maxilla and mandible, temporomandibular joint disorders, and other conditions, the teeth remain central to the diagnostic process \citep{whaites2013essentials, silva2020study, pinheiro2021numbering}. Because of the significance of the teeth and to expedite communication, dentists usually do not identify the teeth by their complete anatomical designations but rather by numbers defined by a system called \textit{Fédération Dentaire Internationale} (FDI) notation \citep{pinheiro2021numbering}. This notation consists of a two-digit system where the numbers can determine the tooth quadrant and type, and is illustrated in Fig. \ref{fig:fdi-system}. We also added a custom color code for each tooth to be used later to show some qualitative results.

Dental reports comprise samples of unstructured data pairs: images and text. Currently, these texts constitute the most extensive and valuable database of dental conditions derived from panoramic radiographs, as they comprehensively describe the entire mouth \citep{jing2017automatic}. The images, in particular, are challenging to interpret due to the vast amount of visual information they contain. Nowadays, computer vision (CV) researchers have been addressing these challenges with relative success through supervised machine-learning techniques. These strategies rely on labeled data, which are manually annotated, restraining their scalability. Text manipulation also presents its challenges, which have previously hindered researchers from utilizing text reports in their work with panoramic radiographs. These challenges arise from the intricacies of natural languages, where nuances, idioms, and contextual meanings can significantly affect interpretation \citep{neveol2018clinical}. Recently, with the advent of large language models (LLMs), the application of artificial intelligence (AI) to text has proven successful. These LLMs are trained in a semi-supervised manner and considerably ease the natural language processing (NLP) tasks \citep{brown2020language}. Consequently, the extraction of dental conditions proved to be feasible yet unexplored by the literature of computational dentistry.

The advances in NLP due to LLMs have turned CV into the most significant bottleneck, primarily because it relies heavily on labeled data. Consequently, no prior work in this domain has conducted its experiments on a dataset of a considerable magnitude. Motivated by this gap, a framework for the classification of thirteen dental conditions in panoramic radiographs is proposed in the current article. The proposed holistic approach incorporates a semi-supervised learning paradigm to address the scarcity of labeled data, a common issue in the medical field. Additionally, a novel auto-labeler strategy that leverages LLMs to extract dental conditions from textual reports is explored.

\subsection{Related Work}

Recent studies in the area of dental radiographic analysis have leveraged machine learning to address various diagnostic challenges, predominantly relying on supervised learning. For instance, \cite{ekert2019deep} developed a custom seven-layer neural network to detect apical lesions in teeth using manually cropped panoramic radiographs, achieving an Area under the ROC Curve (AUC) of 0.85. Similarly, \cite{fukuda2020evaluation} employed the DetectNet CNN architecture to identify vertical root fractures in a dataset of 330 panoramic radiographs, reporting a promising F1-score of 0.83.

In the odontogenic cystic lesions (OCLs) domain, \cite{lee2020diagnosis} used GoogLeNet Inception-v3 to distinguish between odontogenic keratocyst, dentigerous cysts, and periapical cysts. The study utilized 2,126 high-quality images and found the method particularly effective when trained with cone-beam computed tomography (CBCT) images. Building on this, \cite{kwon2020automatic} employed YOLOv3 to identify four types of dental cysts using 1,282 labeled panoramic radiographs, achieving high accuracy despite the limited dataset size.

To enhance the diagnosis of various dental conditions, \cite{chen2021dental} developed a CNN-based auxiliary system for detecting lesions in 2,900 periapical radiographs, focusing on different disease categories and severity levels. The results were notably promising for severe lesions. Additionally, \cite{yuksel2021dental} introduced DENTECT, a three-stage pipeline for classifying five dental conditions using panoramic radiographs, demonstrating performance comparable to dental clinicians.

Segmentation of dental conditions has also been explored. \cite{khan2021automated} evaluated U-Net-based architectures for identifying caries, alveolar bone recession, and interradicular radiolucencies in 206 periapical radiographs, suggesting that custom-built solutions could improve performance. Similarly, \cite{vinayahalingam2021automated} used a Mask R-CNN to segment and classify teeth and five dental conditions in 2,000 radiographs, achieving high F1-scores and underscoring the potential of deep learning in clinical diagnostics.

Several studies have compared different neural network architectures for dental condition classification. \cite{liu2023recognition} evaluated ResNet-50, VGG-16, InceptionV3, and DenseNet-121 on 188 periapical radiographs, with DenseNet-121 achieving the highest accuracy of 99.5\%. \cite{bonfanti2022evaluation} assessed Denti.AI™ for detecting five dental conditions, finding it effective for implants, crowns, and metal fillings but limited in classifying resin-based restorations.

Innovative approaches include DiagnoCat \citep{amasya2024development}, which used Mask R-CNN and Cascade R-CNN to detect periodontal bone loss in 6,000 panoramic radiographs, achieving accurate results in comparison with clinician assessments. Additionally, \cite{rashidi2023autonomous} focused on predicting eight future dental treatments using YOLOv7, showing high accuracy and potential for clinical application.

Lastly, \cite{gao2024ai} introduced YOLO-DENTAL for detecting and classifying dental conditions in 413 periapical radiographs, achieving a notable mean average precision (mAP) of 86.81\%. Similarly, \cite{tassoker2024performance} used YOLOv5 to detect idiopathic osteosclerosis in 175 panoramic radiographs, reporting high detection accuracy despite dataset challenges. These studies collectively highlight the evolving landscape of machine learning applications in dental radiographic analysis.

\begin{table*}[]

\caption{Comparison of the current study with others regarding dataset, task, learning paradigm, proposed solution, and investigated classes.}
\label{tab:relatedwork}

{\fontsize{2.5}{4}\selectfont
\begin{landscape}
\begin{adjustbox}{angle=90}

\begin{tabular}{@{}lcccccccccl@{}}
\toprule
\textbf{Reference} & \textbf{Radiograph}                                                 & \textbf{\begin{tabular}[c]{@{}c@{}}\# Radiographs\end{tabular}} & \textbf{Detection} & \textbf{\begin{tabular}[c]{@{}c@{}}Auto-\\ labeling\end{tabular}} & \textbf{Supervision} & \textbf{\begin{tabular}[c]{@{}c@{}}Semi-\\ supervision\end{tabular}} & \textbf{\begin{tabular}[c]{@{}c@{}}Self-\\ supervision\end{tabular}} & \textbf{\begin{tabular}[c]{@{}c@{}}Architecture\\ or framework\end{tabular}}               & \textbf{\# Classes} & \textbf{Classes}                                                                                                                                                                                            \\ \midrule
\cite{ekert2019deep}               & Panoramic & 85                                                                             &              &                  & \checkmark              &                     &                     & Custom                                                                                     & 2                   & \begin{tabular}[l]{@{}c@{}}Two levels of\\ apical lesions\end{tabular}                                                                                                                                      \\ \hline
\cite{fukuda2020evaluation}               & Panoramic                                                           & 300                                                                            & \checkmark            &                  & \checkmark              &                     &                     & \begin{tabular}[l]{@{}l@{}}DetectNet\\ version V5\end{tabular}                             & 1                   & Vertical root fracture                                                                                                                                                                                      \\ \hline
\cite{lee2020diagnosis}               & \begin{tabular}[l]{@{}c@{}}CBCT\\ and panoramic\end{tabular}        & 2,126                                                                           &              &                  & \checkmark              &                     &                     & \begin{tabular}[l]{@{}l@{}}GoogLeNet\\ Inception‐v3\end{tabular}                           & 3                   & \begin{tabular}[l]{@{}l@{}}Odontogenic keratocysts\\Dentigerous cysts\\ Periapical cysts\end{tabular}                                                                                                 \\ \hline
\cite{kwon2020automatic}               & Panoramic                                                           & 1,282                                                                           & \checkmark            &                  & \checkmark              &                     &                     & YOLOv3                                                                                     & 4                   & \begin{tabular}[l]{@{}l@{}}Dentigerous cysts\\Periapical cysts \\ Odontogenic keratocysts \\Ameloblastomas\end{tabular}                                                                               \\ \hline
\cite{chen2021dental}             & Periapical                                                          & 2,900                                                                           & \checkmark            &                  & \checkmark              &                     &                     & Faster R-CNN                                                                               & 9                   & \begin{tabular}[l]{@{}l@{}}Severities for decay (3)\\ Periapical periodontitis (3)\\ Periodontitis (3)\end{tabular}                                                                                \\ \hline
\cite{yuksel2021dental}               & Panoramic                                                           & 1,005                                                               & \checkmark            &                  & \checkmark              &                     &                     & \begin{tabular}[l]{@{}c@{}}Custom framework\\ DENTECT\end{tabular}                         & 5                   & \begin{tabular}[l]{@{}l@{}}Periapical lesion therapy\\ Fillings \\ Root canal treatment \\ Surgical extraction \\ Conventional extraction\end{tabular}                                                  \\ \hline

\cite{khan2021automated}               & Periapical                                                          & 206                                                                            & \checkmark            &                  & \checkmark              &                     &                     & U-Net                                                                                      & 3                   & \begin{tabular}[l]{@{}l@{}}Caries \\ Alveolar bone recession \\ Interradicular radiolucencies\end{tabular}                                                                                               \\ \hline
\cite{vinayahalingam2021automated}               & Panoramic                                                           & 2,000                                                                           & \checkmark            &                  & \checkmark              &                     &                     & \begin{tabular}[l]{@{}l@{}}Mask R-CNN\\ backboned\\ by Resnet-50\end{tabular}              & 6                   & \begin{tabular}[l]{@{}l@{}}Crowns \\ Fillings \\ Root canal fillings \\ Implants \\ Root remnants\end{tabular}                                                                                           \\ \hline
\cite{liu2023recognition}               & Periapical                                                          & 188                                                                            &              &                  & \checkmark              &                     &                     & \begin{tabular}[l]{@{}l@{}}ResNet-50;\\ Vgg-16;\\ InceptionV3;\\ DenseNet-121\end{tabular} & 3                   & \begin{tabular}[l]{@{}l@{}}Caries \\ Periapical peridontites \\ Periapical cyst\end{tabular}                                                                                                             \\ \hline
\cite{bonfanti2022evaluation}               & Panoramic                                                           & 300                                                                            & \checkmark            &                  & \checkmark              &                     &                     & Faster R-CNN                                                                               & 5                   & \begin{tabular}[l]{@{}l@{}}Metal restorations \\ Resin-based restorations \\ Endodontic treatment \\ Crowns \\ Implants\end{tabular}                                                                     \\ \hline
\cite{amasya2024development}               & Panoramic                                                           & 6000                                                                            & \checkmark            &                  & \checkmark              &                     &                     & \begin{tabular}[l]{@{}l@{}}Mask R-CNN\\ and Faster R-CNN\end{tabular}                      & 1                   & Periodontal bone loss detection                                                                                                                                                                             \\ \hline
\cite{rashidi2023autonomous}               & Panoramic                                                           & 1,025                                                                           & \checkmark            &                  & \checkmark              &                     &                     & YOLOv7                                                                                     & 8                   & \begin{tabular}[l]{@{}l@{}}Restoration \\ Root canal treatment and filling\\ Crown \\ Conventional extraction \\ Bridge \\ Implant\\ Root canal treatment and filling\\ Surgical extraction\end{tabular} \\ \hline
\cite{gao2024ai}             & Periapical                                                          & 1,567                                                                           & \checkmark            &                  & \checkmark              &                     &                     & YOLOv7                                                                                     & 4                   & \begin{tabular}[l]{@{}l@{}}Dental caries \\ Dental defects \\ Periapical lesions \\ Coronal restorations\end{tabular}                                                                                    \\ \hline
\cite{tassoker2024performance}               & Panoramic                                                           & 175                                                                            & \checkmark            &                  & \checkmark              &                     &                     & YOLOv5                                                                                     & 1                   & Idiopathic osteosclerosis                                                                                                                                                                                   \\ \hline
\textbf{Ours}      & \textbf{Panoramic}                                                  & \textbf{16,824}                                                    & \textbf{\checkmark}   & \textbf{\checkmark}        & \textbf{\checkmark}     & \textbf{\checkmark}          & \textbf{\checkmark}          & \textbf{Custom}             & \textbf{13}         & \textbf{Several} \
\end{tabular}
\end{adjustbox}
\end{landscape}
}
\end{table*}

\subsection{Contributions}

Table \ref{tab:relatedwork} compares the current study with other research in dental radiographic automation, highlighting common limitations. Most studies, except for one using a dataset of 6,000 images \citep{amasya2024development}, relied on datasets with 2,900 or fewer samples \citep{chen2021dental, lee2020diagnosis, vinayahalingam2021automated}. Such small datasets can compromise model generalizability. Additionally, many studies excluded challenging cases, limiting the practical applicability of their findings \citep{fukuda2020evaluation, lee2020diagnosis, chen2021dental, liu2023recognition, amasya2024development, gao2024ai}. The focus on a narrow range of target classes \citep{ekert2019deep, fukuda2020evaluation, lee2020diagnosis, khan2021automated, liu2023recognition, amasya2024development} further restricts the comprehensiveness of these models.

Technically, most studies relied on supervised learning \citep{rashidi2023autonomous, tassoker2024performance, amasya2024development, bonfanti2022evaluation}, which requires extensive labeled data, which is a significant limitation. There was also a trend towards using off-the-shelf solutions \citep{lee2020diagnosis, kwon2020automatic, vinayahalingam2021automated, liu2023recognition, rashidi2023autonomous}.

This study introduces a novel framework for diagnosing various dental conditions from panoramic radiographs, leveraging the largest dataset in the field (16,824 images). A semi-supervised approach, combining human-in-the-loop (HITL) strategy \citep{silva2023boosting} and masked autoencoders (MAE) \cite{he2022masked}, is used to enhance performance and reliability. The proposed framework involves creating tooth crops from annotated and predicted teeth on radiographs, with an auto-labeler using LLMs to extract dental conditions from textual reports and map them to corresponding teeth using the FDI numbering system. This approach covers 13 dental conditions and includes a statistical agreement analysis to validate the results.

\section{Classifying multiple dental conditions through a semi-supervised framework}

\begin{figure}[t!]
    \centering
    \includegraphics[width=0.42\textwidth]{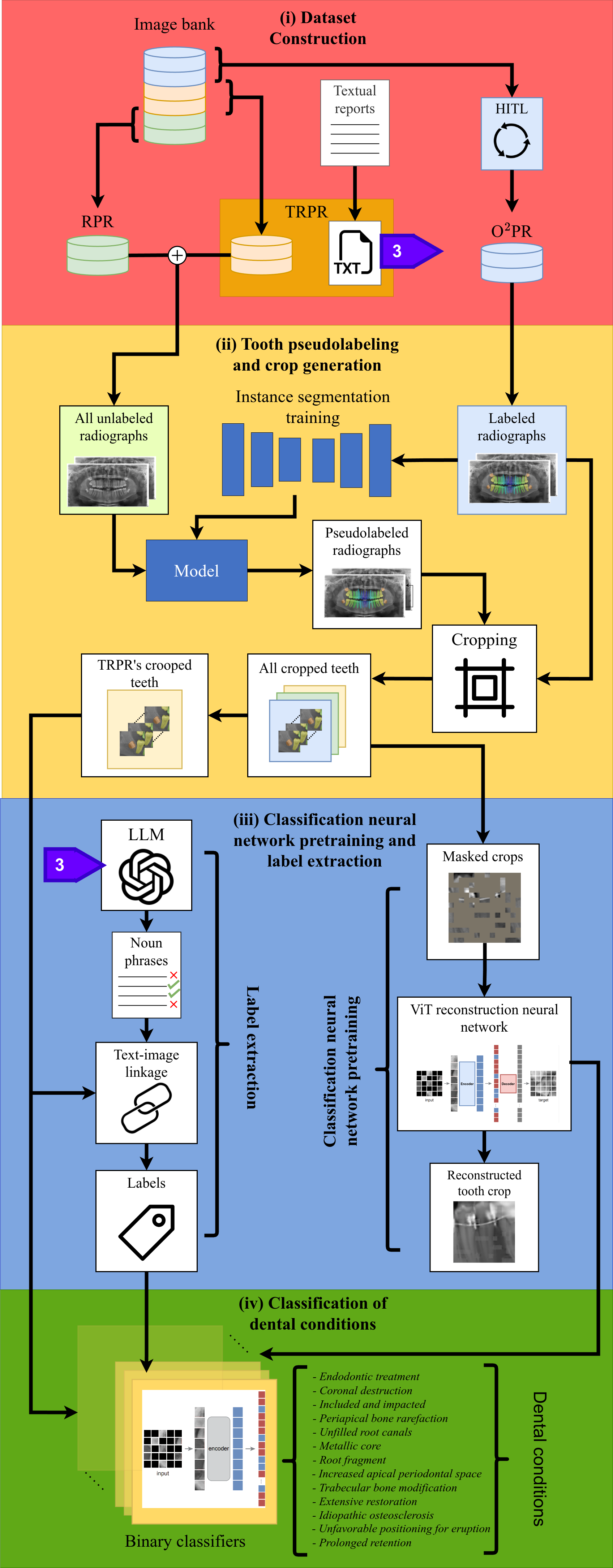}
    \caption{Semi-supervised framework for classifying dental conditions. (i) \textbf{Dataset construction}: Combines Textual Report Panoramic Radiographs (TRPR), Raw Panoramic Radiographs (RPR), and \text{O$^2$PR}. (ii) \textbf{Tooth pseudolabeling and crop generation}: Uses an instance segmentation neural network to generate tooth pseudolabels on unlabeled radiographs, creating tooth crops. (iii) \textbf{Classification network pretraining and label extraction}: Cropped teeth and text information pretrain a model via MAE, extracting noun phrases with a large language model (LLM). Text-image linkage generates labels for binary classifiers. (iv) \textbf{Classification of dental conditions}: Trains a binary classifier for each dental condition.}
    
    \label{fig:pipeline}
\end{figure}

The proposed framework consists of four major steps: (i) Dataset construction, (ii) Tooth pseudolabeling and crop generation, (iii) Classification network pretraining and label extraction, and (iv) Dental conditions classification. These steps are depicted in Fig. \ref{fig:pipeline} and described in the following.

\subsection{Dataset construction}

 Two distinct groups of datasets were built, as detailed in Tables \ref{tab:datasets-one} and \ref{tab:datasets-two} \footnote{The National Commission for Research Ethics (CONEP) and the Research Ethics Committee (CEP) authorized the use of the radiographs in research under the report number 646.050/2014.\\}. The first group comprises a set of panoramic radiographs in total dimensions. The second group, derived from the first, comprises tooth crops for training binary classification models.

The first group's preparation step is illustrated in Fig. \ref{fig:pipeline}(i). This group of radiographs was sourced from an image bank with 16,824 samples, all acquired using the same imaging device \citep{silva2023boosting}. These images were gathered into three distinct, non-overlapping datasets. The first dataset encompasses 4,795 unlabeled images retained in their original, raw format. Therefore, it is designated here as the Raw Panoramic Radiographs (RPR) dataset. The second dataset, the O$^2$PR, was released by \cite{silva2023boosting} and built under the human-in-the-loop (HITL) concept. It includes 4,000 radiograph images along with tooth instance segmentation labels and numbering. Finally, the third dataset, devoid of tooth segmentation labels, comprises 8,029 panoramic radiographs and the corresponding textual reports of two radiologists working independently. It is designated here as the Textual Report Panoramic Radiographs (TRPR) dataset. The main characteristics of these datasets are presented in Table \ref{tab:datasets-one}.

\begin{table*}[t]
\centering
{\tiny
\centering
\caption{Feature summary of the first group of datasets, whose images were obtained from the image bank in their full dimensions. These features include image counts, tooth instance segmentation labels, textual report availability, tooth pseudolabeling, image dimensions, and dataset splitting for training an instance segmentation model.}
\label{tab:datasets-one}
\begin{tabular}{|ccccccc|}
\hline
\multicolumn{1}{|c|}{\multirow{3}{*}{Dataset}} & \multicolumn{1}{c|}{\multirow{3}{*}{\# Images}} & \multicolumn{1}{c|}{\multirow{3}{*}{\begin{tabular}[c]{@{}c@{}}Inst. Segm.\\ Labels\end{tabular}}} & \multicolumn{1}{c|}{\multirow{3}{*}{\begin{tabular}[c]{@{}c@{}}Textual\\ Reports\end{tabular}}} & \multicolumn{1}{c|}{\multirow{3}{*}{Pseudolabeling}} & \multicolumn{1}{c|}{\multirow{3}{*}{\begin{tabular}[c]{@{}c@{}}Mode\\ Dimensions\end{tabular}}} & \multirow{3}{*}{\begin{tabular}[c]{@{}c@{}}Inst. Segm.\\Training Data\end{tabular}} \\
\multicolumn{1}{|c|}{}                         & \multicolumn{1}{c|}{}                           & \multicolumn{1}{c|}{}                                                                              & \multicolumn{1}{c|}{}                                                                           & \multicolumn{1}{c|}{}                                 & \multicolumn{1}{c|}{}                                                                          &                        \\
\multicolumn{1}{|c|}{}                         & \multicolumn{1}{c|}{}                           & \multicolumn{1}{c|}{}                                                                              & \multicolumn{1}{c|}{}                                                                           & \multicolumn{1}{c|}{}                                 & \multicolumn{1}{c|}{}                                                                          &                        \\ \hline
\multicolumn{1}{|c|}{RPR}                      & \multicolumn{1}{c|}{4,795}                      & \multicolumn{1}{l|}{}                                                                              & \multicolumn{1}{l|}{}                                                                           & \multicolumn{1}{c|}{\checkmark}        & \multicolumn{1}{c|}{\multirow{3}{*}{2,440$\times$1,292}}                                         & None                   \\ \cline{1-5} \cline{7-7} 
\multicolumn{1}{|c|}{O$^2$PR}                  & \multicolumn{1}{c|}{4,000}                      & \multicolumn{1}{c|}{\checkmark}                                                     & \multicolumn{1}{l|}{}                                                                           & \multicolumn{1}{l|}{}                                 & \multicolumn{1}{c|}{}                                                                          & All                    \\ \cline{1-5} \cline{7-7} 
\multicolumn{1}{|c|}{TRPR}                     & \multicolumn{1}{c|}{8,029}                      & \multicolumn{1}{l|}{}                                                                              & \multicolumn{1}{c|}{\checkmark}                                                  & \multicolumn{1}{c|}{\checkmark}        & \multicolumn{1}{c|}{}                                                                          & None                   \\ \hline
\end{tabular}
}
\end{table*}

\begin{table*}[t]
\centering
{\fontsize{4.5}{7.5}\selectfont
\centering
\caption{Feature summary of the second group of datasets, termed \textbf{Crops}, whose images were used to pretrain and train binary classifiers for tooth conditions. These characteristics include the image dimensions, source dataset, crop counts, textual reports availability, pretraining usage, and dataset splitting to train and test the binary classifiers. All the images are tooth crops sourced from the first group of datasets.}
\label{tab:datasets-two}
\begin{tabular}{|c|c|c|c|c|c|c|c|c|}
\hline
Dataset & \begin{tabular}[c]{@{}c@{}}Crop\\ Dimensions\end{tabular} & \begin{tabular}[c]{@{}c@{}}Source\\ Dataset\end{tabular} & \# Crops & \begin{tabular}[c]{@{}c@{}}Textual\\ Reports\end{tabular} & Pretraining & Train & Validation & Test \\ \hline
\multirow{3}{*}{\begin{tabular}[c]{@{}c@{}}Less\\ context\end{tabular}} & \multirow{3}{*}{224$\times$224} & \multirow{2}{*}{RPR} & \multirow{2}{*}{132,497} & \multirow{4}{*}{} & \multirow{4}{*}{All} & \multirow{4}{*}{None} & \multirow{4}{*}{None} & \multirow{4}{*}{None} \\
 &  &  &  &  &  &  &  &  \\ \cline{3-4}
 &  & \multirow{2}{*}{O2PR} & \multirow{2}{*}{112,842} &  &  &  &  &  \\ \cline{1-2}
\multirow{3}{*}{\begin{tabular}[c]{@{}c@{}}More\\ context\end{tabular}} & \multirow{3}{*}{\begin{tabular}[c]{@{}c@{}}380$\times$380\\ to 224$\times$224\end{tabular}} &  &  &  &  &  &  &  \\ \cline{3-9} 
 &  & \multirow{2}{*}{TRPR} & \multirow{2}{*}{213,395} & \multirow{2}{*}{\checkmark} & \multirow{2}{*}{Train only (70\%)} & \multirow{2}{*}{70\%} & \multirow{2}{*}{15\%} & \multirow{2}{*}{15\%} \\
 &  &  &  &  &  &  &  &  \\ \hline
\end{tabular}
}
\end{table*}

The images had widths ranging from 2,272 to 2,692 and heights ranging from 1,292 to 1,304. The most common size was 2,440 $\times$ 1,292, accounting for 79\% of the images. The textual reports were preprocessed from the original \texttt{.odt} format and converted to the \texttt{.txt} format for consistency and easier manipulation. The final \texttt{.txt} files had the format exemplified in Table \ref{tab:samples} \footnote{For this article, all textual reports were translated from their original language (Portuguese) to English.}: A two-digit number followed by a colon and the report line. In Table \ref{tab:samples}, the teeth are highlighted in bold according to the FDI system described in Fig. \ref{fig:fdi-system}, to emphasize their significance for the radiograph screening. In the used textual reports, as commonly done, the digits of the textual reports denote a tooth through the FDI notation, while other numbers are written out in full.

The second group of datasets is synthesized from the previous datasets (RPR, O$^2$PR, and TRPR). It comprises tooth crop images, which will be detailed in the following section.

\begin{table}[t]
\centering
\caption{Sample of a panoramic radiograph preprocessed report of the TRPR dataset. It is highlighted in bold the mentioning of the teeth according to the FDI system described in Fig \ref{fig:fdi-system} to reveal their importance.}
\begin{tabular}{@{}cp{6cm}@{}}

\toprule
\multicolumn{1}{c}{Topic Number} & \multicolumn{1}{c}{Report line}                                                                                   \\ \midrule
01:          & Anatomical modification in the right and left mandible condyle.                        \\
02:          & Missing teeth: \textbf{18}, \textbf{28} and \textbf{48}.                                                                 \\
03:          & Teeth \textbf{13} and \textbf{38} included and impacted.                                                        \\
04:          & Tooth \textbf{36} and \textbf{37}: endodontic treatment. Partially filled root canals.                          \\
05:          & Mild bone loss in the region of the present teeth.                                            \\
06:          & Modification of the bone trabeculation in the region of tooth \textbf{48} compatible with a bone scar. \\
07:          & Calcification of the right and left stylohyoid ligament complex.                             \\ \bottomrule
\end{tabular}
\label{tab:samples}
\end{table}

\subsection{Tooth pseudo labeling and crop generation}

This step serves two purposes, as illustrated in Fig. \ref{fig:pipeline}(ii): (i) to automatically generate pseudo labels for the teeth in the radiographs of the RPR and TRPR datasets using an instance segmentation network, as they do not contain instance segmentation labels; and (ii) to create tooth crops from the tooth labels and pseudo labels, which are later used to train and test the classification neural networks. These procedures allow consistent, uniformly sized image crops over all datasets around each tooth. The standardization is crucial for subsequent steps, as the classification neural networks are trained on fixed-size inputs. Notably, the instance segmentation network's output is achieved by training it with ``pseudolabels''.

An instance segmentation neural network was trained on all 4,000 labeled images of the O$^2$PR dataset. The hybrid task cascade (HTC) architecture \citep{chen2019hybrid} was selected because it was the best model in the benchmark conducted by \cite{silva2023boosting}. HTC ensures more accurate object boundaries and improved detection results by leveraging information from tasks like semantic segmentation. Using this network,  the teeth of all the remaining unlabeled radiographs from the RPR (4,795) and the train set of TRPR (8,029) datasets were segmented. Finally, 4,000 labeled radiographs and 12,824 pseudo labeled radiographs were obtained.

\begin{figure}[]
    \centering
    \includegraphics[width=\linewidth]{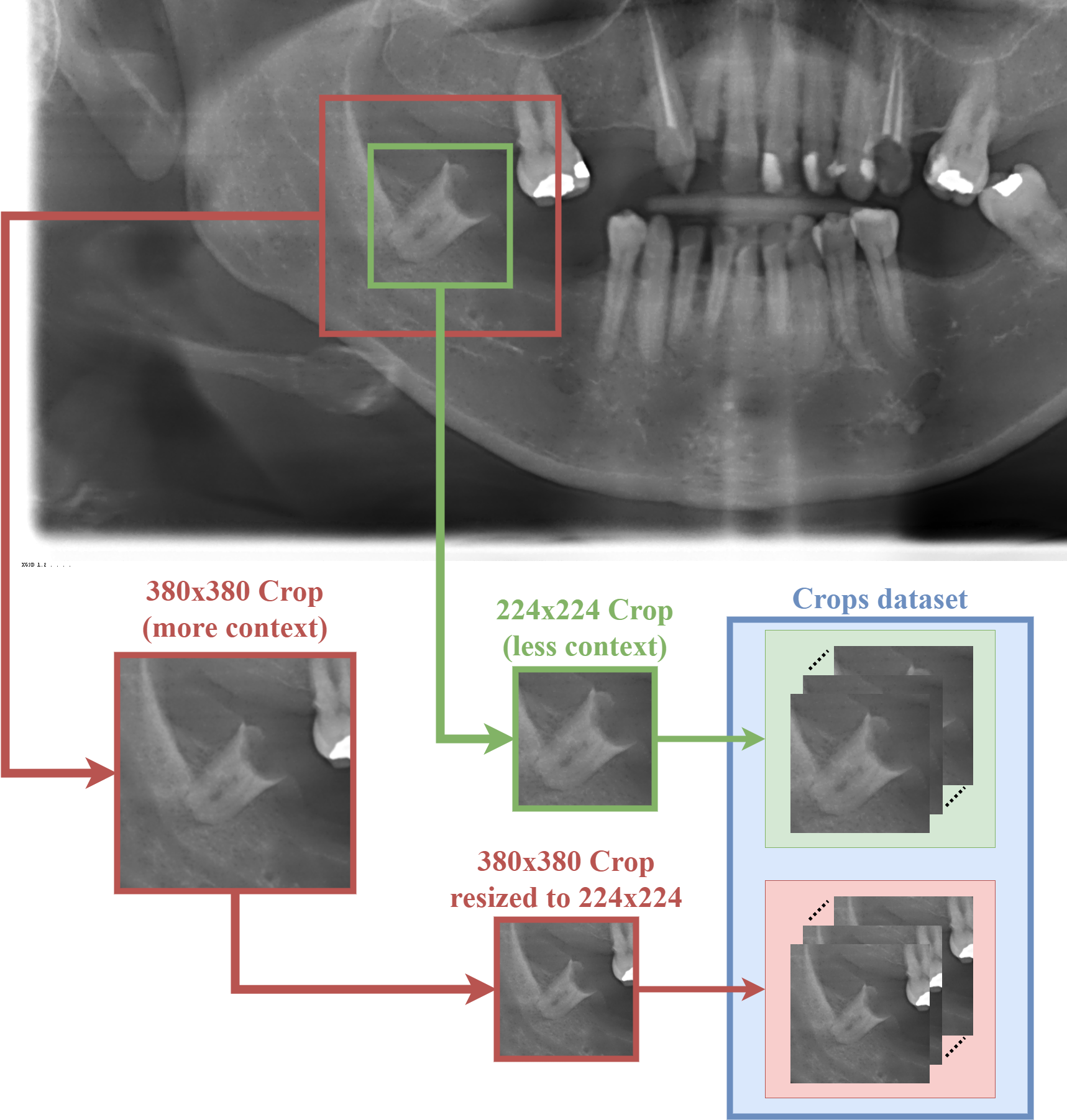}

    \caption{Two tooth crop variants used in this study: The first, termed the ``less context'' crop, was taken from a panoramic radiograph of a tooth and measures 224$\times$224 pixels. The second, termed the ``more context'' crop, was resized to 224$\times$224 pixels from an original crop of size 380$\times$380 pixels. These two sets comprise the Crops dataset.}

    \label{fig:380to380crops}
\end{figure}

After training, our HTC instance segmentation neural network was used to create two distinct datasets of tooth crops from all labeled and pseudolabeled radiographs. The primary variant spanned $224 \times 224$ centered around each tooth. This tooth crop type has the advantage of being more focused on the teeth but the disadvantage of having less context of the tooth surroundings, possibly excluding tooth parts. This dataset is termed ``less context'' crops. To address the context gap, a second crop category, termed ``more context'' crops, began with a broader $380 \times 380$ area, which was then resized to $224 \times 224$ to comply with the requirements of the employed neural network architecture used for classification. After this procedure, approximately 460,000 crops were obtained for each configuration. Fig. \ref{fig:380to380crops} illustrates the cropping procedure, while Table \ref{tab:datasets-two} compiles the features of these datasets, termed \textbf{Crops}.

\subsection{Classification network pretraining and label extraction}
\label{subsection:class-labels}
The next step is illustrated in Fig. \ref{fig:pipeline}(iii). This stage consists of two distinct processes that can be executed independently: the neural network pretraining and label extraction (via auto-labeler). The neural network pretraining was designed to enhance the performance of classification models for the final tasks, leveraging the unlabeled data to learn significant features. Besides the baselines discussed in Section \ref{subsection:Neural-network-pretraining}, this process was run independently twice, once for each set of the tooth Crops dataset. Tooth crops of $224 \times 224$ pixel dimensions were used to meet the requirements of the chosen classification model: the Vision Transformer (ViT) \citep{dosovitskiy2020image}. The ViT was selected for the used pretraining strategy because of its superior performance in benchmarks \citep{khan2022transformers}, where it achieved state-of-the-art results by exploiting transformers, and due to its convenience for employing the Mask AutoEncoder (MAE) strategy.
 
\begin{figure}[t]
    \centering
    \includegraphics[width=\linewidth]{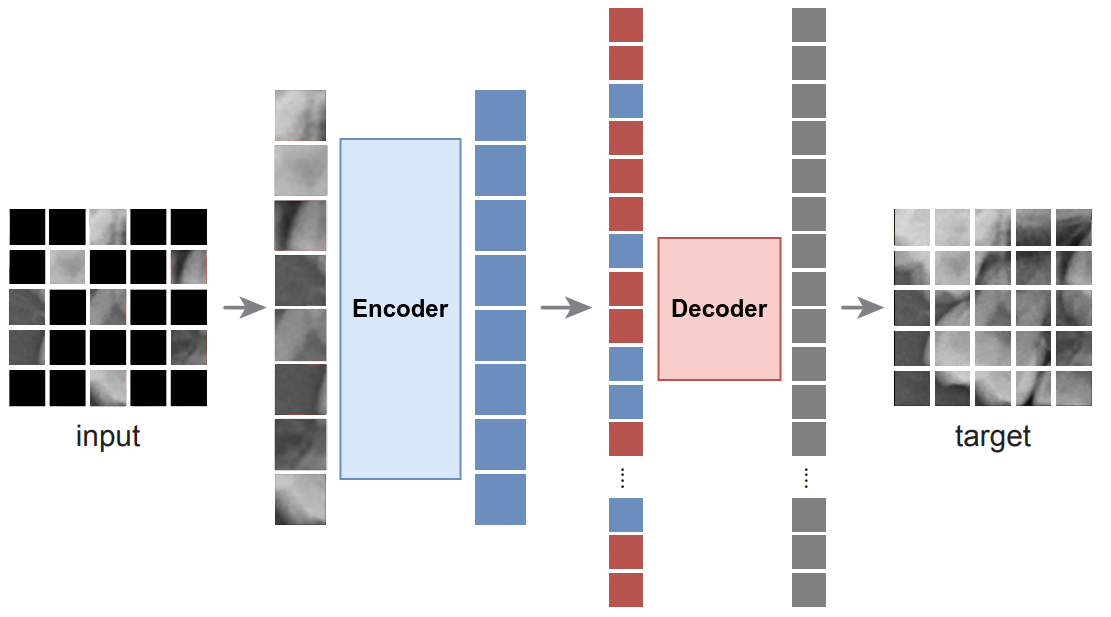}
    \caption{Illustration of a MAE: Selected patches from an input image are obscured, and the remaining visible patches are processed through an encoder. The obscured patches are subsequently reconstructed using a decoder from the latent space representations.}
    \label{fig:mae-pipeline}
\end{figure}

The current study used MAE, a variant of deep-learning autoencoders trained by deliberately masking out portions of the input data, as a pretraining strategy. This ``masking'' approach challenges the network to reconstruct the original data, including the intentionally obscured (masked) patches, resulting in a more robust latent representation. Benefiting from transfer learning principles, MAEs can leverage pretrained models to further enhance their performance and generalization capabilities. Fig. \ref{fig:mae-pipeline} illustrates an MAE within the context of the ViT architecture. Given an input image segmented into a grid with some sections obscured, the encoder compresses this partial image into a compact representation. These encoded data are then processed by the decoder, aiming to regenerate the full image. In the selected configuration, the decoder has fewer parameters than the encoder to focus on efficient reconstruction \citep{he2022masked}. After completing the pretraining phase, the decoder is discarded, emphasizing its role in learning robust image representations without contributing to the final task performance.

The auto-labeler is the second process, executed in parallel with the pretraining procedure. The labels here represent dental conditions; therefore, only the crops from the TRPR dataset were used, as TRPR is the only dataset containing textual reports. This process aimed to extract the noun phrases to create the labels later used as ground truth for the classification neural networks. A noun phrase is a word or group of words with a noun as its head or main word. Noun phrases can function in a sentence as a subject, an object, or a complement. They can be single nouns or more complex structures with modifiers and related words. In the proposed framework, extracting noun phrases is necessary because all dental conditions are noun phrases, although not all noun phrases are dental conditions.

The state-of-the-art GPT-4 \citep{openai2023gpt4} was used to automate and expedite noun phrase extraction. The adopted methodology is centered on prompt engineering -- a technique where specific inputs are crafted to elicit desired outputs from language models. Guided by this approach, the following prompt was formulated to be executed on the textual reports:

\begin{quote}
``You are an excellent English teacher who indicates for each item (sentences starting with two digits) of a textual report the noun phrases through vertical topics."
\end{quote}

With the input above, for instance, the noun phrases extracted from sentence ``03'' in the report shown in Table \ref{tab:samples} were:

\begin{itemize}
    \setlength\itemsep{0em}
    \item Tooth 36
    \item 37
    \item Endodontic treatment
    \item Partially filled root canals
\end{itemize}

However, in the lists of noun phrases, the teeth are not directly connected to their conditions. A linkage procedure was used to solve this issue. In this procedure, all teeth were associated with all present conditions in the sentence. For instance, using the linkage procedure, tooth 36 and tooth 37 both have the conditions of endodontic treatment and partially filled root canals. This example is illustrated in the following line:
\newline
\begin{center}
\begin{tikzpicture}
    \hspace{-0.5em}%
    \node (formula) [align=left] {\underline{Tooth 36} and \underline{37}: \underline{endodontic treatment}. \underline{Partiall}y\underline{ filled root canals}.};
    \draw[-latex,black] ($(formula.north west)+(0.8,0)$) arc
    [
        start angle=160,
        end angle=20,
        x radius=1.8cm,
        y radius =0.5cm
    ];

    \draw[-latex,black] ($(formula.north west)+(0.8,0)$) arc
    [
        start angle=160,
        end angle=15,
        x radius=3.3cm,
        y radius =0.75cm
    ] ;

    \draw[-latex,black] ($(formula.north west)+(2.25,0)$) arc
    [
        start angle=160,
        end angle=20,
        x radius=1.cm,
        y radius =0.25cm
    ] ;

    \draw[-latex,black] ($(formula.north west)+(2.25,0)$) arc
    [
        start angle=160,
        end angle=15,
        x radius=2.5cm,
        y radius=0.5cm
    ] ;

\end{tikzpicture}
\end{center}

The linkage process proves effective in the context of dental reports, which are organized by specific conditions. Radiologists are more favorable to note the presence of conditions rather than the absence. This tendency is illustrated in the case of the patient mentioned in Table \ref{tab:samples}, sentence ``03''. For example, it is less common for radiologists to document each tooth lacking conditions, such as `Teeth 17, 16, 15, 14, 13, ... without endodontic treatment or unfilled root canals', focusing instead on those with notable conditions. 

Sentences detailing the presence or absence of teeth were excluded from the analysis, as illustrated in sentence ``02'' of Table \ref{tab:samples}. This decision was based on the understanding that this task is better conducted using object detection or instance segmentation techniques specifically designed for identifying and delineating objects within images \citep{silva2020study, pinheiro2021numbering}. To filter out these sentences, straightforward regular expressions were implemented.

\subsection{Dental conditions classification}

In Fig. \ref{fig:pipeline}(iv), the focus is on training the ViT networks (binary classifiers). Initially, a baseline model without pre-taining was used. Pretrained weights from MAEs were also explored to broaden the proposed approach. Specifically,  weights from MAEs pretrained on two datasets: The widely recognized ImageNet dataset and the previously constructed Crops dataset were used.

The Matthews correlation coefficient (MCC) metric \citep{matthews1975comparison} was used for the proposed network evaluation. The MCC takes into account true positives (TP), false positives (FP), and false negatives (FN), including the frequently overlooked true negatives (TN):

\begin{equation}
    \label{eq:mcc}
    \text{MCC} = \frac{\text{TP} \times \text{TN} - \text{FP} \times \text{FN}}{\sqrt{(\text{TP} + \text{FP})(\text{TP} + \text{FN})(\text{TN} + \text{FP})(\text{TN} + \text{FN})}} \, .
\end{equation}

The main properties of MCC include: the denominator acts as a normalization factor, bounding MCC values between -1 and 1 (one); a score of -1 indicates entirely incorrect predictions, while 1 (one) signifies perfect predictions; an MCC of 0 (zero) implies predictions equal to random guessing; MCC treats TP, TN, FP, and FN symmetrically, an important feature when these values bear similar implications; MCC is preferable to the cases of imbalanced datasets as it does not give highly optimist results in opposition to other metrics, such as accuracy.

\section{Experimental analysis}

\subsection{Training for tooth pseudo labeling}

The main goal of this stage was training an instance segmentation neural network, specifically the HTC. Following the methodology outlined by \cite{silva2023boosting}, the trained HTC used a ResNeXt neural network as its backbone with 101 layers and a cardinality of 64 \citep{xie2017aggregated}. The initial weights of this network were derived from the training on the ImageNet dataset to leverage the transfer of the learning technique later. The training data comprised 4,000 images from the O$^2$PR dataset. No data was allocated for testing, emphasizing that the goal was not efficiency measurement but tooth crop generation for subsequent phases.

\begin{figure}
    \centering
    \begin{subfigure}{\linewidth}
        \centering
        \includegraphics[width=0.95\linewidth]{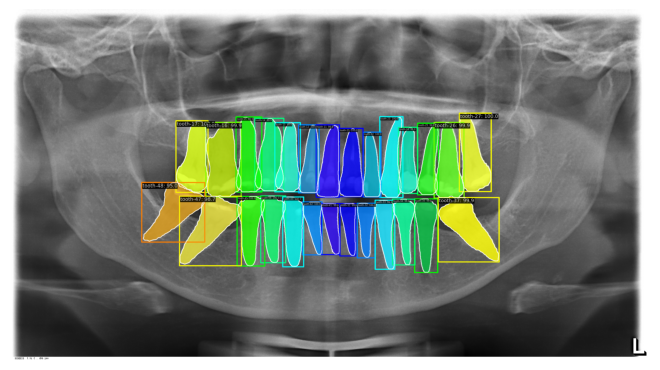}
    \caption{Instance segmentation results of a panoramic radiograph of an adult's mouth.}
    \label{fig:qualitative-result-a}
    \end{subfigure}
    \begin{subfigure}{\linewidth}
        \centering
        \includegraphics[width=0.95\linewidth]{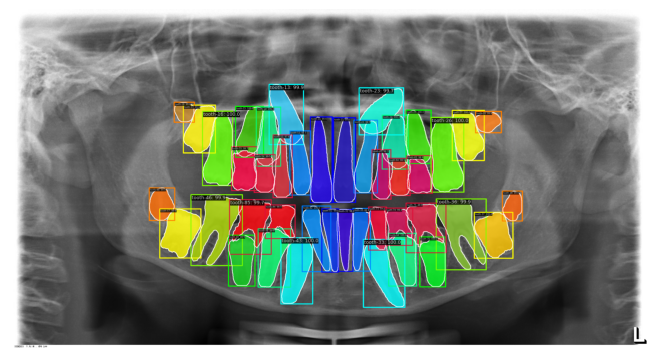}
    \caption{Instance segmentation results of a panoramic radiograph of a child's mouth.}
    \label{fig:qualitative-result-b}
    \end{subfigure}
    \caption{Qualitative results of the trained instance segmentation neural network, using the color code introduced in Fig. \ref{fig:fdi-system}.}
    \label{fig:qualitative-result}
\end{figure}

Data augmentation was purely horizontal flips, carefully changing the labels of the teeth from the right side to the left and vice versa. To optimize network performance, the radiographs were cropped from their prevailing 2,440 $\times$ 1,292 pixels to 1,876 $\times$ 1,036, removing 159 pixels from the top, which resulted in more focused teeth. The batch size was 1 (one), and the optimizer was stochastic gradient descent, with a learning rate of 0.0015, momentum of 0.9, and no weight decay. The threshold value for tooth detection was 0.5. The network was trained for 20 epochs in an NVIDIA GeForce GTX TITAN X. After training the neural network; it was applied to the 12,824 unlabeled images from the RPR and TRPR datasets. This allowed one to create tooth crops for all the radiographs in the used image dataset. Fig. \ref{fig:qualitative-result} shows two samples of the instance segmentation results, in an adult’s and a child’s mouth, using the color code introduced in Fig. \ref{fig:fdi-system}, demonstrating its promising performance.

\subsection{Neural network pretraining}
\label{subsection:Neural-network-pretraining}

The MAE technique was exploited to pretrain neural networks for subsequent transfer learning to final classification networks for each dental condition.

All the available data of tooth crops was used for pretraining the ViTs, reserving some images for validation and testing purposes (see Table \ref{tab:datasets-two}). The experiment encompassed three scenarios: the first employed a baseline network devoid of pretraining, the second involved networks pretrained on the ImageNet dataset, and the third used custom-generated tooth crops. Each scenario was executed twice, accommodating both crop configurations, culminating in six distinct experimental setups.

\begin{itemize}
    \setlength\itemsep{0em}
    \item Less context crops without pretraining
    \item Less context crops pretrained on ImageNet dataset
    \item Less context crops pretrained on Crops dataset
    \item More context crops without pretraining
    \item More context crops pretrained on ImageNet dataset
    \item More context crops pretrained on Crops dataset
\end{itemize}

\begin{figure}[t]
    \centering
    \includegraphics[width=0.8\linewidth]{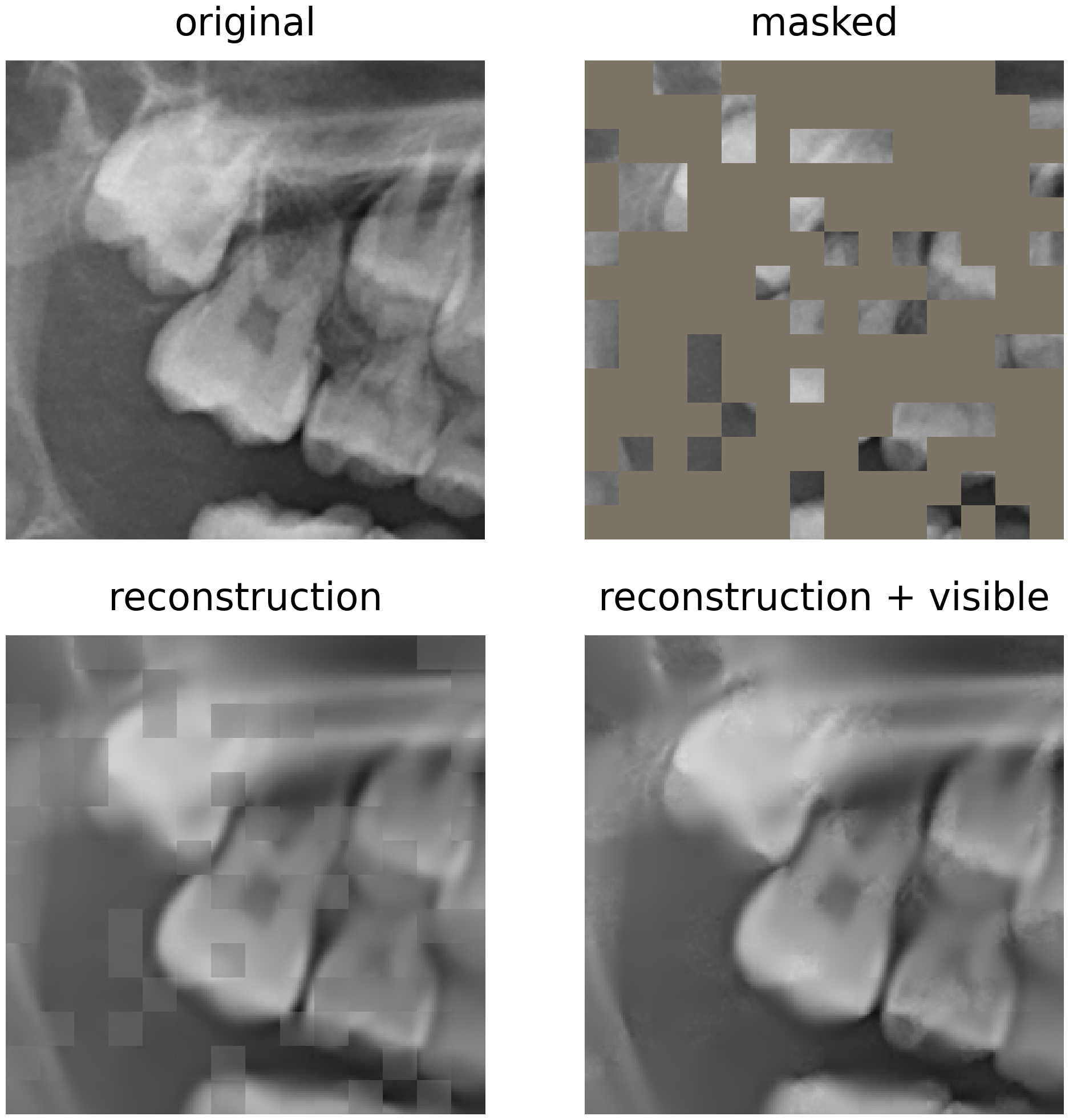}
    \caption{Reconstruction sample from a pretrained neural network using MAE as a pretraining strategy, showcasing the efficacy of MAE in enhancing the network's reconstructive capabilities.}
    \label{fig:reconstrion}
\end{figure}

In pretraining scenarios, data augmentation techniques used horizontal flip and random resized crop, with scales ranging from 0.2 to 1. The batch size was 512, and the optimizer was AdamW with a learning rate of $9.5 \times 10^{-4}$, betas of 0.9 and 0.95, and no weight decay. The network was trained for 800 epochs with a linear warm-up in the first 40 epochs. The hardware used for training was eight NVIDIA A100 of 80 GB. The depicted sample in Fig. \ref{fig:reconstrion} showcases the considerable qualitative success of the reconstruction outcomes from the pretraining configuration using tooth crops.

\subsection {Label extraction}

\begin{figure*}[t]
    \centering
    \includegraphics[width=\linewidth]{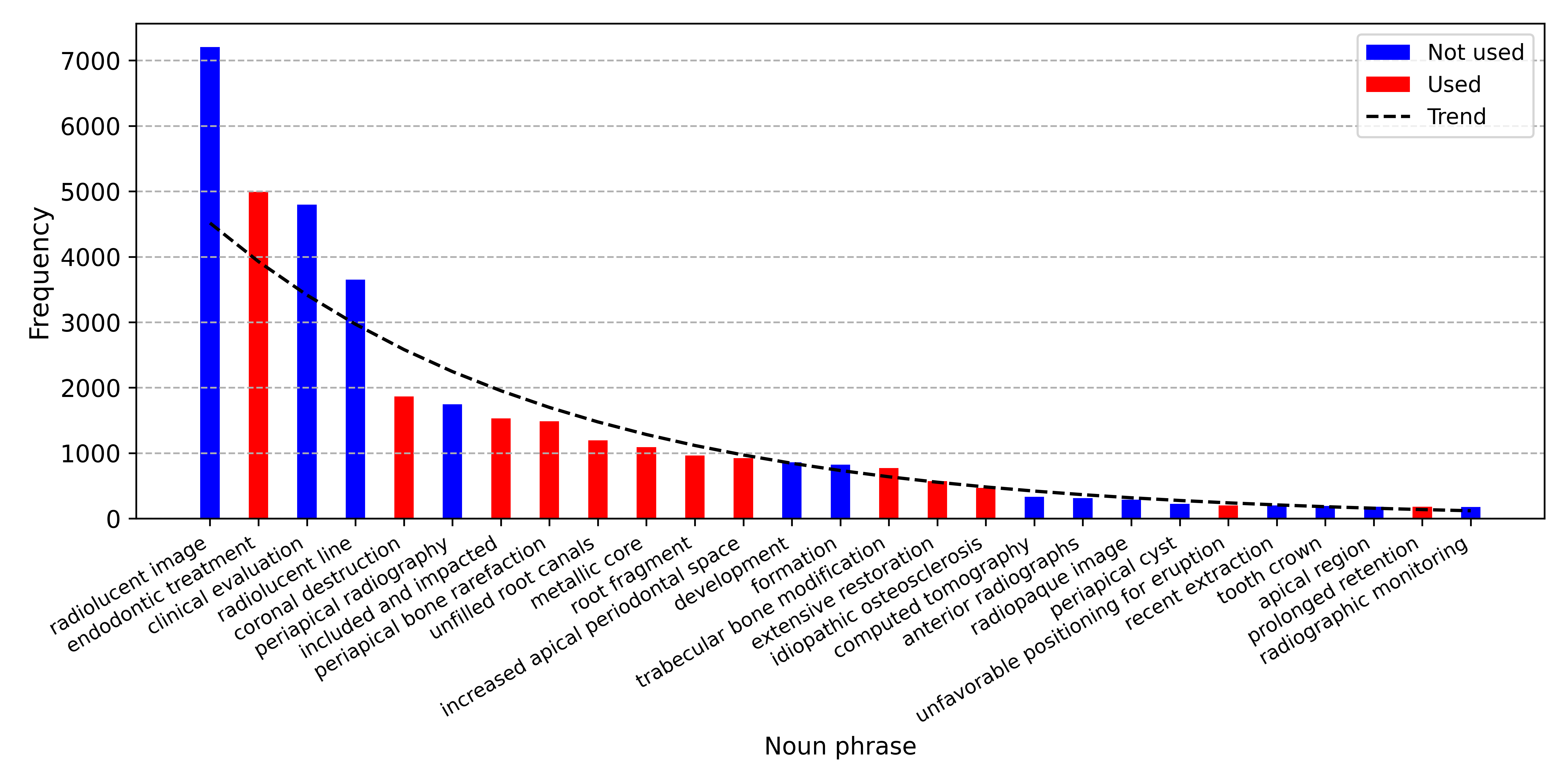}
    \caption{Bar chart of the 27 most common noun phrases, showing their frequency and trends, and illustrating their long tail distribution. (Noun phrases identified as dental conditions are highlighted with red bars, while those not selected are represented with blue bars.)}
    \label{fig:frequency-noun-phrases}
\end{figure*}

In this phase, OpenAI's LLM GPT-4 was used to streamline and expedite the extraction of noun phrases from textual reports of the TRPR dataset. For the current study, the frequency of all noun phrases was gauged, and only those with occurrences higher than 150 were considered. This threshold was chosen arbitrarily, believing it to represent the minimum necessary for a network to learn effectively. Afterward, similar phrases, such as ``unfilled root canal'' with its plural form ``unfilled root canals'', were manually grouped. As not all noun phrases are dental conditions, the selection was refined through manual filtering. For example, ``endodontic treatment'' is a dental condition, whereas ``clinical assessment'' is not and, therefore, was  excluded from the analysis. Fig. \ref{fig:frequency-noun-phrases} displays a bar chart depicting the 27 most common noun phrases, their frequency, and trends, evincing their long tail distribution. Noun phrases identified as dental conditions are marked with red bars, whereas those not selected are shown with blue bars.

In the end, the descriptions of the selected dental conditions, each assigned a unique numerical index and ordered by their frequency of occurrence (indicated in parentheses), were:

\begin{enumerate}
\item Endodontic treatment (4,994) - a procedure that treats infections inside the tooth, typically involving the removal of the pulp and nerves, followed by the filling and sealing of the pulp chamber and root canals.
\item Coronal destruction (1,866) - Damage or decay to the crown portion of the tooth.
\item Included and impacted (1,532) - Teeth trapped within the jawbone or gums and cannot erupt naturally.
\item Periapical bone rarefaction (1,486) - A reduction or loss of bone density around the apex of a tooth root, often due to inflammation or infection.
\item Unfilled root canals (1,194) - Root canals that have not been filled or sealed after an endodontic procedure.
\item Metallic core (1,091) - A metal post used to support a restoration or crown, especially in a tooth undergoing endodontic treatment.
\item Root fragment (964) - A piece or portion of a tooth root left behind, typically after tooth extraction or breakage.
\item Increased apical periodontal space (922) - Enlargement of the space around the tooth root's apex, which may indicate an inflammatory response.
\item Trabecular bone modification (773) - Changes in the spongy part of the bone, which can be indicative of disease or other conditions.
\item Extensive restoration (573) - Large dental fillings or excessive material used in a dental restoration.
\item Idiopathic osteosclerosis (470) - A localized increase in bone density without a known cause.
\item Unfavorable positioning for eruption (200) - The positioning of a tooth that hinders its natural eruption process.
\item Prolonged retention (181) - The extended presence of a tooth or dental element beyond its normal duration, often referring to baby teeth that don't fall out on time.
\end{enumerate}

\begin{figure*}[ht!p]
    \centering
    
    \begin{subfigure}{0.12\linewidth}
        \includegraphics[width=\linewidth]{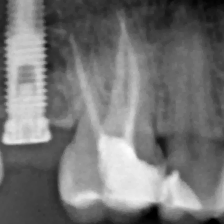}
        \caption{\centering Endodontic treatment. \textcolor{white}{....} }
    \end{subfigure}
    \hfill
    \begin{subfigure}{0.12\linewidth}
        \includegraphics[width=\linewidth]{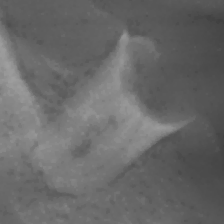}
        \caption{\centering { }{ }Coronal{ }{ } destruction.}
    \end{subfigure}
    \hfill
    \begin{subfigure}{0.12\linewidth}
       \includegraphics[width=\linewidth]{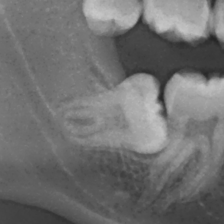}
        \caption{\centering Included and impacted.}
    \end{subfigure}
    \hfill
    \begin{subfigure}{0.12\linewidth}
        \includegraphics[width=\linewidth]{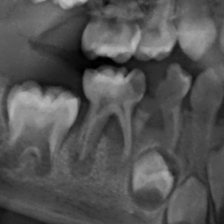}
        \caption{\centering Periapical bone rarefaction.}
    \end{subfigure}
    \hfill
    \begin{subfigure}{0.12\linewidth}
        \includegraphics[width=\linewidth]{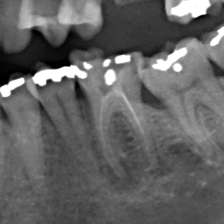}
        \caption{\centering \textcolor{white}{.}Unfilled root canals. \textcolor{white}{....}}

    \end{subfigure}
    \hfill
    \begin{subfigure}{0.12\linewidth}

        \includegraphics[width=\linewidth]{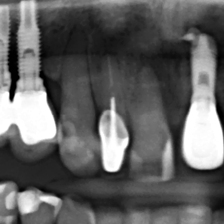}
        \caption{\centering { }Metallic core. \textcolor{white}{..........}}
    \end{subfigure}
    \hfill
    \begin{subfigure}{0.12\linewidth}
        \includegraphics[width=\linewidth]{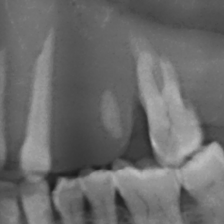}
    \caption{\centering { }{ }{ }{ }{ }Root fragment. \textcolor{white}{......}}
    \end{subfigure}

    \vspace{0.25cm}

    \begin{subfigure}{0.12\linewidth}
        \includegraphics[width=\linewidth]{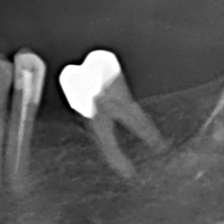}
        \caption{\centering Increased apical periodontal space.}
    \end{subfigure}
    \hfill
    \begin{subfigure}{0.12\linewidth}
        \includegraphics[width=\linewidth]{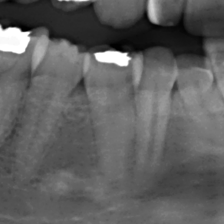}
        \caption{\centering Trabecular bone modification. \textcolor{white}{......}}
    \end{subfigure}
    \hfill
    \begin{subfigure}{0.12\linewidth}
        \includegraphics[width=\linewidth]{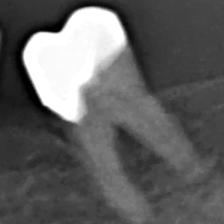}
        \caption{\centering Extensive restoration.  \textcolor{white}{......}  \textcolor{white}{......}  \textcolor{white}{......}}
    \end{subfigure}
    \hfill
    \begin{subfigure}{0.12\linewidth}
        \includegraphics[width=\linewidth]{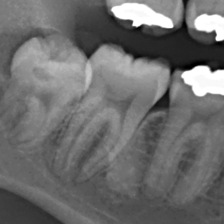}
        \caption{\centering Idiopathic osteosclerosis.  \textcolor{white}{......}  \textcolor{white}{......}}
    \end{subfigure}
    \hfill
    \begin{subfigure}{0.12\linewidth}
        \includegraphics[width=\linewidth]{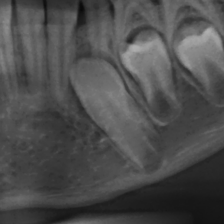}
        \caption{\centering Unfavorable positioning for eruption.}
    \end{subfigure}
    \hfill
    \begin{subfigure}{0.12\linewidth}
        \includegraphics[width=\linewidth]{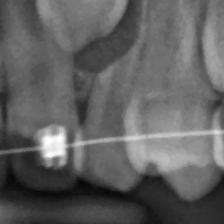}
        \caption{\centering Prolonged retention.  \textcolor{white}{......}  \textcolor{white}{......}  \textcolor{white}{......}}
    \end{subfigure}

    \caption{Dental conditions considered in this study. (They were selected according to their frequency in the textual reports.)}
    \label{fig:each-con-sample}
\end{figure*}

Fig. \ref{fig:each-con-sample} displays examples of each condition. Upon determining the conditions to be evaluated, the adopted linkage process, which associates every tooth mentioned in a sentence with all the dental conditions stated in that sentence, was applied as described in Section \ref{subsection:class-labels}.

\subsection{Classification neural network training}

The TRPR dataset was split into train (70\%), validation (15\%), and test (15\%) subsets for training and evaluation (see Table \ref{tab:datasets-one}. The tooth crops were $224 \times 224$ without resizing (less context), or the $224 \times 224$ resized from $380 \times 380$ (more context) crops. Data augmentation techniques used were horizontal flip 50\% of the time, 10 degrees random rotation, and color jitter with parameters 0.2 for brightness, 0.2 contrast, and 0.2 saturation. The positive classes were oversampled by a factor of 10 due to their insufficient representation. The batch size was 64, and the optimizer was AdamW with a base learning rate of $10^{-3}$, betas of 0.9 and 0.95, and no weight decay. The network was trained for 50 epochs with no linear warm-up. The hardware used for training was eight NVIDIA A100 of 80 GB. Finally, the loss was binary cross entropy (BCE) plus MCC loss were calculated:

\begin{equation}
\text{BCE}(y, \hat{y}) = -\frac{1}{N} \sum_{i=1}^{N} \left(y_i \cdot \log(\hat{y}_i) + (1 - y_i) \cdot \log(1 - \hat{y}_i)\right) \, ,
\end{equation}
where $N$ is the number of samples, $y$ is a vector of the target labels, and the $\hat{y}$ is the vector of the predicted probabilities.

The MCC loss is given by $1 - \text{MCC}$, where a small value was added on the denominator of Eq. \ref{eq:mcc} to avoid division by 0 (zero).

The final loss is
\begin{equation}
    \text{Loss} = \alpha \cdot \text{BCE}(y, \hat{y}) + (1- \alpha) \cdot (1 - \text{MCC}(y, \hat{y})) \, .
\end{equation}
Here, $\alpha = 0.5$.

\subsection{Results and discussions}

Table \ref{tab:mcc-results} showcases the primary numerical MCC results. It presents, for each tooth condition (\textbf{Label}), the positive class sample size frequency (\textbf{Freq.}), the \textbf{Validation} outcomes for each pretraining configuration, the maximum MCC value on validation datasets (\textbf{Max Val.}), and the \textbf{Test} results. One can conclude from the validation average values that the no-pretraining configurations, indicated in Tables \ref{tab:mcc-results} and \ref{tab:mcc-epochs} by the column \textbf{None}, had the worst results both on the less-context and more-context tooth crop scenarios. In contrast, the pretraining from the tooth crop dataset, indicated in the tables by \textbf{Crops}, had the best average results in both cases. Pretraining with tooth crop data outperformed \textbf{ImageNet} pretraining, on average, by 6.73 percentage points (p.p.) and 1.14 (p.p.) in the less-context and more-context tooth crops, respectively. Despite containing approximately 460,000 images--far fewer than the ImageNet dataset's more than 17 million--pretraining with tooth crop data proved more efficient due to the field-oriented data context. The faster convergence of tooth crop pretraining configurations demonstrates its efficiency. Table \ref{tab:mcc-epochs} shows that tooth crop pretraining configurations perform more optimally in fewer epochs than those pretraining with ImageNet.

\begin{table*}[t]
\centering
\caption{Results based on the MCC values from the validation and test sets indicate that pretraining with the ImageNet and Crops dataset was beneficial.}
\label{tab:mcc-results}
{\tiny
\begin{tabular}{|cc|cccccc|c|c|}
\hline
\multicolumn{1}{|c|}{\multirow{4}{*}{\textbf{Label}}} & \multirow{4}{*}{\textbf{Freq.}} & \multicolumn{6}{c|}{\textbf{Validation}}      
                              
& \multirow{4}{*}{\textbf{Max Val.}} & \multirow{4}{*}{\textbf{Test}} \\ \cline{3-8}
\multicolumn{1}{|c|}{}                                &                                 & \multicolumn{3}{c|}{\multirow{2}{*}{\textbf{\begin{tabular}[c]{@{}c@{}}224 × 224 crops\\ (less context)\end{tabular}}}} & \multicolumn{3}{c|}{\multirow{2}{*}{\textbf{\begin{tabular}[c]{@{}c@{}}380 × 380 crops\\ (more context)\end{tabular}}}} &                                   &                                \\
\multicolumn{1}{|c|}{}                                &                                 & \multicolumn{3}{c|}{}                                                                                                   & \multicolumn{3}{c|}{}                                                                                                                        &                                   &                                \\ \cline{3-8}
\multicolumn{1}{|c|}{}                                &                                 & \multicolumn{1}{c|}{\textbf{None}}   & \multicolumn{1}{c|}{\textbf{ImageNet}}   & \multicolumn{1}{c|}{\textbf{Crops}}   & \multicolumn{1}{c|}{\textbf{None}}               & \multicolumn{1}{c|}{\textbf{ImageNet}}               & \textbf{Crops}               &                                   &                                \\ \hline
\multicolumn{1}{|c|}{1}                               & 4,994                            & 0.864                                & 0.903                                    & \multicolumn{1}{c|}{\textbf{0.904}}   & 0.827                                            & 0.847                                                & 0.846                              & 0.904                    & 0.865                          \\
\multicolumn{1}{|c|}{2}                               & 1,866                            & 0.421                                & 0.668                                    & \multicolumn{1}{c|}{\textbf{0.714}}   & 0.300                                            & 0.663                                                & 0.675                              & 0.714                    & 0.658                          \\
\multicolumn{1}{|c|}{3}                               & 1,532                            & 0.681                                & 0.767                                    & \multicolumn{1}{c|}{0.740}            & 0.649                                            & 0.776                                                & \textbf{0.790}                     & 0.790                    & 0.683                          \\
\multicolumn{1}{|c|}{4}                               & 1,486                            & 0.455                                & 0.589                                    & \multicolumn{1}{c|}{\textbf{0.598}}   & 0.487                                            & 0.498                                                & 0.523                              & 0.598                    & 0.397                          \\
\multicolumn{1}{|c|}{5}                               & 1,194                            & 0.264                                & 0.454                                    & \multicolumn{1}{c|}{0.445}            & 0.455                                            & \textbf{0.611}                                       & 0.595                              & 0.611                    & 0.436                          \\
\multicolumn{1}{|c|}{6}                               & 1,091                            & 0.653                                & 0.677                                    & \multicolumn{1}{c|}{0.695}            & 0.150                                            & 0.711                                                & \textbf{0.750}                     & 0.750                    & 0.632                          \\
\multicolumn{1}{|c|}{7}                               & 964                             & 0.318                                & 0.532                                    & \multicolumn{1}{c|}{0.510}            & 0.167                                            & \textbf{0.728}                                       & 0.668                              & 0.728                    & 0.583                          \\
\multicolumn{1}{|c|}{8}                               & 922                             & 0.142                                & 0.275                                    & \multicolumn{1}{c|}{0.270}            & 0.394                                            & 0.399                                                & \textbf{0.405}                     & 0.405                    & 0.327                          \\
\multicolumn{1}{|c|}{9}                               & 773                             & 0.000                                & 0.301                                    & \multicolumn{1}{c|}{0.309}            & 0.649                                            & \textbf{0.506}                                       & 0.458                              & 0.649                    & 0.218                          \\
\multicolumn{1}{|c|}{10}                              & 573                             & 0.000                                & \textbf{0.385}                           & \multicolumn{1}{c|}{0.286}            & 0.000                                            & 0.284                                                & 0.314                              & 0.385                    & 0.252                          \\
\multicolumn{1}{|c|}{11}                              & 470                             & 0.000                                & 0.182                                    & \multicolumn{1}{c|}{0.182}            & 0.000                                            & \textbf{0.424}                                       & 0.414                              & 0.424                    & 0.347                          \\
\multicolumn{1}{|c|}{12}                              & 200                             & 0.299                                & 0.336                                    & \multicolumn{1}{c|}{0.420}            & 0.302                                            & 0.430                                                & \textbf{0.456}                     & 0.456                    & 0.353                          \\
\multicolumn{1}{|c|}{13}                              & 181                             & 0.240                                & 0.577                                    & \multicolumn{1}{c|}{\textbf{0.666}}   & 0.211                                            & 0.386                                                & 0.545                              & 0.666                    & 0.426                          \\ \hline
\multicolumn{2}{|c|}{\textbf{Average}}                                                  & 0.334                                & 0.511                                    & \multicolumn{1}{c|}{0.519}                                 & 0.353
& 0.559                                                & 0.572                              & 0.622                             & 0.475                          \\ \hline
\end{tabular}
}
\end{table*}

\begin{table*}[t]
\centering
\caption{Analysis of epoch convergences (values in the table), based on the highest MCC value on the validation sets.}
\label{tab:mcc-epochs}
{\scriptsize
\begin{tabular}{|cc|cccccc|}
\hline
\multicolumn{1}{|c|}{\multirow{4}{*}{\textbf{Label}}} & \multirow{4}{*}{\textbf{Freq.}} & \multicolumn{6}{c|}{\textbf{Validation}}                                                                                                                                                                                                                               \\ \cline{3-8} 
\multicolumn{1}{|c|}{}                                &                                 & \multicolumn{3}{c|}{\multirow{2}{*}{\textbf{\begin{tabular}[c]{@{}c@{}}224 × 224 crops\\ (less context)\end{tabular}}}} & \multicolumn{3}{c|}{\multirow{2}{*}{\textbf{\begin{tabular}[c]{@{}c@{}}380 × 380 crops\\ (more context)\end{tabular}}}} \\
\multicolumn{1}{|c|}{}                                &                                 & \multicolumn{3}{c|}{}                                                                                                   & \multicolumn{3}{c|}{}                                                                                                                        \\ \cline{3-8} 
\multicolumn{1}{|c|}{}                                &                                 & \multicolumn{1}{c|}{\textbf{None}}   & \multicolumn{1}{c|}{\textbf{ImageNet}}   & \multicolumn{1}{c|}{\textbf{Crops}}   & \multicolumn{1}{c|}{\textbf{None}}               & \multicolumn{1}{c|}{\textbf{ImageNet}}               & \textbf{Crops}               \\ \hline
\multicolumn{1}{|c|}{1}                               & 4,994                            & 44                                   & 38                                       & \multicolumn{1}{c|}{11}               & 40                                               & 7                                                    & 7                                  \\
\multicolumn{1}{|c|}{2}                               & 1,866                            & 36                                   & 14                                       & \multicolumn{1}{c|}{23}               & 35                                               & 8                                                    & 13                                 \\
\multicolumn{1}{|c|}{3}                               & 1,532                            & 42                                   & 23                                       & \multicolumn{1}{c|}{26}               & 43                                               & 15                                                   & 30                                 \\
\multicolumn{1}{|c|}{4}                               & 1,486                            & 37                                   & 14                                       & \multicolumn{1}{c|}{13}               & 34                                               & 9                                                    & 10                                 \\
\multicolumn{1}{|c|}{5}                               & 1,194                            & 38                                   & 11                                       & \multicolumn{1}{c|}{11}               & 40                                               & 24                                                   & 11                                 \\
\multicolumn{1}{|c|}{6}                               & 1,091                            & 44                                   & 25                                       & \multicolumn{1}{c|}{21}               & 21                                               & 8                                                    & 11                                 \\
\multicolumn{1}{|c|}{7}                               & 964                             & 32                                   & 25                                       & \multicolumn{1}{c|}{11}               & 34                                               & 24                                                   & 7                                  \\
\multicolumn{1}{|c|}{8}                               & 922                             & 37                                   & 3                                        & \multicolumn{1}{c|}{9}                & 28                                               & 6                                                    & 6                                  \\
\multicolumn{1}{|c|}{9}                               & 773                             & 0                                    & 23                                       & \multicolumn{1}{c|}{6}                & 0                                                & 26                                                   & 17                                 \\
\multicolumn{1}{|c|}{10}                              & 573                             & 0                                    & 5                                        & \multicolumn{1}{c|}{12}               & 18                                               & 19                                                   & 1                                  \\
\multicolumn{1}{|c|}{11}                              & 470                             & 0                                    & 16                                       & \multicolumn{1}{c|}{6}                & 0                                                & 6                                                    & 2                                  \\
\multicolumn{1}{|c|}{12}                              & 200                             & 25                                   & 4                                        & \multicolumn{1}{c|}{19}               & 41                                               & 6                                                    & 3                                  \\
\multicolumn{1}{|c|}{13}                              & 181                             & 38                                   & 8                                        & \multicolumn{1}{c|}{10}               & 27                                               & 18                                                   & 11                                 \\ \hline
\multicolumn{2}{|c|}{\textbf{Average}}                                                  & 29                                   & 16                                       & \multicolumn{1}{c|}{14}               & 28                                               & 14                                                   & 10                                 \\ \hline
\end{tabular}
}
\end{table*}

\begin{figure}[t]
    \centering
    \includegraphics[width=\linewidth]{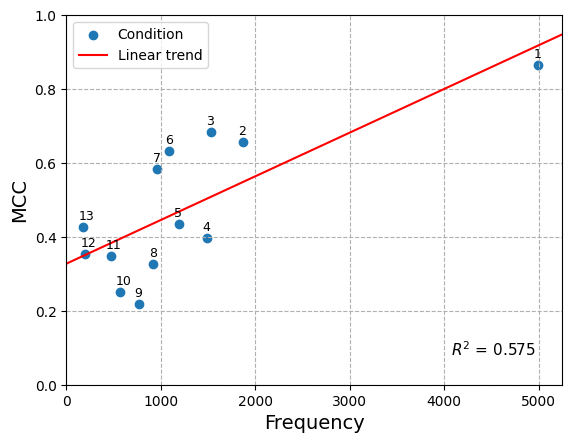}
    \caption{Scatter plot of the MCC results on the test set. (The red line shows the MCC increasing trend according to the frequency of the dental condition in the dataset.)}
    \label{fig:mcc-trend}
\end{figure}

The test set's results in Table \ref{tab:mcc-results} were derived from the top-performing network based on the validation sets. Notably, while the test MCC values exhibit considerable variation, they all exceed 0 (zero), indicating performance better than random guessing. Furthermore, according to the positive sample size, the metrics show a noticeable increasing trend. This trend is illustrated in Fig. \ref{fig:mcc-trend}, a scatter plot of the data, where the trend was computed using a linear function. The linear function $R^2$ reached 0.575. $R^2$ is a statistical measure of how well a mathematical equation represents a set of data. An $R^2$ between 0.5 and 0.7 indicates a substantial fit, meaning the model reliably explains a significant portion of the variance in the data.

While the size of the positive sample contributes to the MCC trend, it does not account for all of it. A deeper understanding of the challenges in classifying different classes offers more insight into how well the network performs. For instance, some conditions are not in the teeth but around them (\textit{e.g.}, in the gum), requiring more image context. These idiosyncrasies are discussed in the following, indicating in parentheses which configuration performed better, whether with less context or more context in the panoramic.

\begin{enumerate}
    \item Endodontic treatment \textbf{(less context)}

    An endodontic treatment appears as white (radiopaque) lines in the tooth canals (refer to Fig. \ref{fig:each-con-sample} (a)). Therefore, an image crop close and centered on the teeth eases the task of identifying this condition. This configuration is the case for the $224 \times 224$ crops. Together with the large amount of positive data, this resulted in a MCC higher than 0.900 on the validation data, while the resize crops from $380 \times  380$ dimensions reached 0.845 MCC.

    \item Coronal destruction \textbf{(less context)}
    
    Coronal destruction appears as darker areas (radiolucencies) because the structure is less dense than a healthy tooth. This decay can be seen as disruptions in the continuous outline of the tooth crown, especially around or underneath existing dental restorations. Therefore, a close, near-the-tooth crop is sufficient for detecting coronal destruction, as depicted in Fig. \ref{fig:each-con-sample} (b). In this case, a maximum of 0.714 was reached from the less-context crops against 0.675 of the more-context one.

    \item Included and impacted \textbf{(more context)}     
    It refers to a tooth that has not erupted into its expected position in the dental arch due to obstruction by another tooth, bone, or soft tissue (Fig. \ref{fig:each-con-sample} (c)). This phenomenon  occurs frequently with wisdom teeth (third molars). Depending on its location, the impacted wisdom tooth can be seen pressing against or tilted towards its neighboring second molar, potentially causing root resorption or displacement. Therefore, a minimal increase in the context of the tooth crop may be beneficial to identify inclusions. The maximum attained results from more-context tooth crops (0.790) were 2.3 (p.p.) higher than the less-context counterpart (0.767).

    \item Periapical bone rarefaction \textbf{(less context)}
    
    Periapical bone rarefaction appears as a darker area (radiolucent) around the tooth's root or at its apex (see Fig. \ref{fig:each-con-sample} (d)). This dark spot indicates bone loss or decreased bone density. The borders of this area can be well-defined or more diffuse, depending on the nature and stage of the condition. Under these circumstances, having a well-focused image around the teeth is better, or even necessary, to better diagnose this condition. Here, the less context reached 0.599 against 0.523.

    \item Unfilled root canals \textbf{(more context)}
    
    On a panoramic radiograph, unfilled root canals within a tooth appear as relatively dark lines or canals within the lighter, radiopaque outline of the tooth structure (refer Fig. \ref{fig:each-con-sample} (e)). These dark lines represent where the dental pulp once was and should generally be filled with endodontic materials if a root canal treatment has been completed. In those conditions, a well-focused image is better for classification. In the current setup, the networks trained with more-context crops attained an MCC of 0.611 and the less-context crops of 0.454.

    \item Metallic core \textbf{(more context)}

    On a panoramic radiograph, a ``metallic core'' within a tooth appears as a highly radiopaque area within the tooth structure, often in the shape of a post or a dense filling (see Fig. \ref{fig:each-con-sample}(f)). It stands out distinctly against the less dense surrounding tooth material and any dental restorations that are not metal-based. However, since it is not a part of the tooth, the crop center excludes the metallic core ``crown''. More context is expected to help the classification in light of this situation. A value of 0.75 for MCC in the more-context configuration and 0.695 in the less-context was reached.

    \item Root fragment \textbf{(more context)}

    On a panoramic radiograph, a ``root fragment'' appears as a radiopaque structure, resembling a part of a tooth's root, as shown in Fig. \ref{fig:each-con-sample}(g). It is usually isolated, without a crown portion, and may be surrounded by a darker area if inflammation or bone resorption is present. The need to screen the tooth's surroundings makes having more context in the image important. Results of 0.728 and 0.532 were achieved in less-context and more-context scenarios, respectively.

\item Increased apical periodontal space \textbf{(more context)}

    On a panoramic radiograph, an increased apical periodontal space, displayed in Fig. \ref{fig:each-con-sample} (h), appears as an enhanced or widened radiolucent area around the tip of the root of a tooth. This dark gap, known as the periodontal ligament space, is usually uniform and thin around the roots of healthy teeth. Bearing this in mind, a closed, well-focused image around the tooth's center may exclude the dental condition, making its diagnosis impossible. Therefore, an image with more context is more beneficial for detecting an increased apical periodontal space. The current study reached a 0.405 MCC in the more context scenario against a 0.275 in the less context (i.e., a performance boost of 47.27\%).

\item Trabecular bone modification \textbf{(more context)}

On a panoramic radiograph, a ``trabecular bone modification'' may appear as changes in the pattern and density of the bone, as shown in Fig. \ref{fig:each-con-sample} (i). Areas with increased density will look whiter, indicating a more solid bone structure, while regions with decreased density will appear darker, suggesting less bone mass. The regular mesh-like pattern of the trabeculae might appear disrupted or altered, which can indicate various dental or bone conditions. These areas appear on the bones surrounding the teeth, not near their center. Therefore, a crop with more context is beneficial for diagnosing trabecular bone modification. In the more-context scenario, we attained 0.506 of MCC; in the less context, we attained 0.309 on the validation datasets.

\item Extensive restoration \textbf{(less context)}

    On a panoramic radiograph, ``extensive restoration'', or ``excess restorative material'', appears as a filling, crown, or other dental work, that extends beyond the natural contours of the tooth, as displayed in Fig. \ref{fig:each-con-sample}(j). Typically used for fillings or crowns, these materials will stand out as they are denser than the tooth and absorb more X-rays. If overfilled, excess material may also be seen beyond the confines of the tooth's normal borders, such as in the interdental spaces or the pulp chamber. Restorative material frequently occurs on the tooth crown, a distance from the tooth center. Therefore, cropping too closely may hinder accurate diagnosis. In the current analysis, the images were not excessively cropped, which allowed for a correct diagnosis with less contextual information. Specifically, an MCC of 0.385 was observed in scenarios with less context compared to 0.314 in scenarios with more context.

\item Idiopathic osteosclerosis \textbf{(more context)}

    On a dental radiograph, ``idiopathic osteosclerosis'' appears as a brighter area due to increased bone density (Fig. \ref{fig:each-con-sample}(k)). It is often seen near the roots of teeth but lacks the characteristic dark border of other lesions. Therefore, the view of the tooth's surroundings could be beneficial for detecting it. The results were 0.424 in the more-context scenario and 0.182 in the less-context one.
    
    \item Unfavorable positioning for eruption \textbf{(more context)}
    
    An ``unfavorable positioning for eruption'' for a tooth appears as a tooth that is misaligned with the normal arch form, often at an abnormal angle or location that suggests it will not erupt into a functional position without intervention (see Fig. \ref{fig:each-con-sample}(m)). This could be a tooth that is tilted, rotated, or horizontally displaced. The context around the tooth is important to verify and confirm if its position is unfavorable for eruption. Indeed, the results in the more-context scenario were 0.456 against 0.420 in the less-context scenario (an increase of 8.57\%).

    \item Prolonged retention \textbf{(less context)}
    
    ``Prolonged retention'' of a tooth is indicated by a tooth that remains in the jaw beyond the typical age of exfoliation without evidence of natural shedding or eruption (Fig. \ref{fig:each-con-sample}(n)). It often appears as a tooth with roots that may be resorbed, situated in the jaw without movement, potentially affecting the positioning of adjacent teeth or the eruption of successor permanent teeth. These deciduous teeth are small and do not require a larger context for diagnosis of prolonged retention. A value of 0.666 was reached in the context scenario against 0.545 in the more context one.
\end{enumerate}

\section{Comparison with dentistry professionals}

The performance evaluation on a large and diverse dataset indicates that the proposed framework has learned, as all MCCs exceeded 0 (zero). However, these values do not provide a desirable comparison to the performance of human professionals when evaluating panoramic radiographs. To make this comparison, annotations made by dentistry professionals were assessed and compared against the results of the classification models' predictions. This was accomplished by inviting five final-year undergraduate students (junior annotators) and five radiologist experts (senior annotators) to label some samples of the same test images used to evaluate the classification models.

In the labeling setup, each participant had to label a cropped panoramic radiograph centered on a specific tooth, similar to the images used for training and evaluating the models. The images in the setup were the same as the ``more-context'' $380 \times 380$ crops, but without resizing to $224 \times 224$, which was previously necessary to meet the model's input requirements. The dentistry professionals had to identify and mark all visible dental conditions in the area of the central tooth in the crop, based on the provided options (all the conditions considered in this study), or mark none, according to their analysis.

A hurdle that needed to be overcome in this procedure was the number of test images to be annotated. According to the estimates, there were about 32,000 test images (please refer to Table \ref{tab:datasets-two}), which would require more than 300 hours of continuous work for each participant to label, making it impractical. A natural alternative was to sample a subset of the test set while maintaining the positive/negative class proportions. Unfortunately, this option also proved unworkable due to the highly imbalanced datasets. For instance, the positive/negative ratio for condition 13 (prolonged retention) is 0.106. In this case, the professionals would need to annotate approximately 1,180 samples to maintain the proportional ratio, with only one being positive. Annotating such a large number of image crops was beyond reasonable feasibility. The issue was overcome by selecting images through a strategy that ensured a minimum number of positive examples and variability. The adopted strategy consisted of selecting 78 samples (six images per condition) using the following pattern: for each condition, two true positives (TP), two false positives (FP), and two false negatives (FN) were selected based on the original models' predictions. According to the labels extracted from the reports, this approach ensured at least four positive samples (the two TPs and the two FNs) and two negative samples (FP). It also provided potential variability due to the FP and FN typically being borderline cases. This final set of images is designated as \textbf{Expert Image Dataset}.

\subsection{Initial assessment}

Table \ref{tab:avg-first-results} shows the results for each professional and the average results for the students and experts, considering the labels from the text reports as the ground truth. The outcomes indicate that the expert group performed moderately better than the students (0.429 vs. 0.455 MCC). The attained MCC by the used models was 0.475 (Table \ref{tab:mcc-results}), which is higher than the scores of both groups, demonstrating strong performance. However, one cannot assert that the models have reached superhuman performance because the scenario is biased in favor of the models. Rather than being trained to detect dental conditions in general, the models were trained to detect conditions as the primary labeler (reports), giving them an advantage over professionals who did not have access to the annotator samples. This issue was mitigated by combining the expert labels, as discussed below.

\begin{table}[t]
\caption{Average MCC for each student and expert (the average value for each group is also included.)}
\centering
\label{tab:avg-first-results}
\begin{tabular}{|c|c|c|c|}
\hline
\rowcolor[HTML]{EFEFEF} 
\textbf{Student} & \textbf{Avg. MCC} & \textbf{Expert} & \textbf{Avg. MCC} \\ \hline
Student 1 & 0.479 & Expert 1 & 0.394 \\ \hline
Student 2 & 0.500 & Expert 2 & 0.589 \\ \hline
Student 3 & 0.360 & Expert 3 & 0.387 \\ \hline
Student 4 & 0.435 & Expert 4 & 0.455 \\ \hline
Student 5 & 0.372 & Expert 5 & 0.452 \\ \hline
\textbf{All Students} & \textbf{0.429} & \textbf{All Experts} & \textbf{0.455} \\ \hline
\end{tabular}
\end{table}

\subsection{Definitive assessment with expert consensus}

The original training, validation, and testing labels were derived from textual reports. Under these conditions, the models trained and validated on these datasets had an advantage over professionals, as with the proposed solution, when benchmarked. To mitigate this bias in the models' performance, the labels provided by the professionals were leveraged to create a new ground truth.

It was assumed that the expert could generate the most accurate labels. Therefore, it was decided that, for the proposed solution and the students, the new ground truth for the Expert Image Dataset would be generated by combining the labels from all experts. This setup not only avoided favoring the proposed solution but also increased robustness and reduced labeling noise, as a majority vote of the annotators created the new labels. For the experts,  a leave-one-out layout was built where, for each of the five rounds, the ground truth was computed from the labels of four experts, and the remaining specialist was evaluated against this new ground truth.


Under this new layout, the results for the models and the average results for the students and experts were 50.8\%, 51.0\%, and 58.9\%, respectively. The detailed results for each participant are included in Table \ref{tab:avg-final-results}. It is possible to draw two conclusions from these numbers. Firstly, the MCC values are considerably higher than those attained on the reports' ground truth. This increase can be attributed to the more robust ground truths that were less noisy, as the combination of the labels of several experts generated them. Furthermore, the similarity between the MCCs attained by the proposed solution and the students (50.82\% vs 51.00\%) led one to conclude that the proposed solution reached the level of a junior professional.

\begin{table}[t]
\caption{Final average MCC of all conditions for each student and expert, including the average value for each group.}
\label{tab:avg-final-results}
\centering
\begin{tabular}{|c|c|c|c|}
\hline
\rowcolor[HTML]{EFEFEF} 
\textbf{Student} & \textbf{Avg. MCC} & \textbf{Expert} & \textbf{Avg. MCC} \\ \hline
Student 1 & 0.527 & Expert 1 & 0.607 \\ \hline
Student 2 & 0.591 & Expert 2 & 0.689 \\ \hline
Student 3 & 0.490 & Expert 3 & 0.499 \\ \hline
Student 4 & 0.516 & Expert 4 & 0.575 \\ \hline
Student 5 & 0.426 & Expert 5 & 0.574 \\ \hline
\textbf{All Students} & \textbf{0.510} & \textbf{All Experts} & \textbf{0.589} \\ \hline
\end{tabular}
\end{table}

\begin{figure*}[t]
    \centering
    \includegraphics[width=\linewidth]{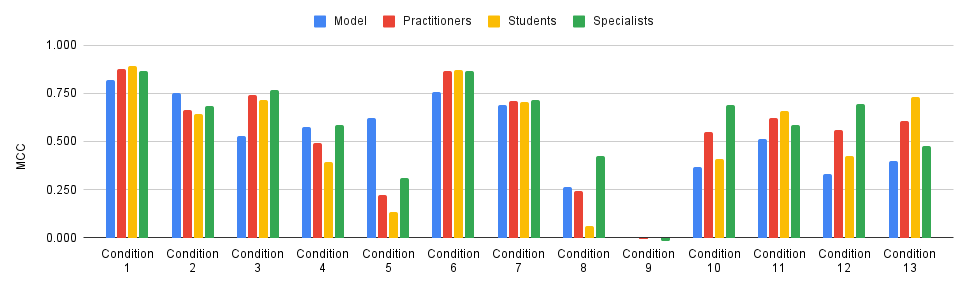}
    \caption{Bar chart that breaks down the definitive assessment results according to condition and professional group. (The results on Condition 9 (trabecular bone modification) were closer to 0 (zero).)}
    \label{fig:statistical-graph-per-dental-and-group-gt}
\end{figure*}

Fig. \ref{fig:statistical-graph-per-dental-and-group-gt} depicts a bar chart that further investigates the results, breaking them down by classes and predictors (models, professionals, students, and experts). One can observe that the proposed solution, when compared to the professionals, demonstrates significantly higher performance in Condition 5 (unfilled root canals) and considerably worse performance in Condition 12 (unfavorable positioning for eruption) and 13 (prolonged retention), the ones of less positive samples. However, what stands out the most is the almost null MCC values for condition 9 (Trabecular bone modification) class. One can hypothesize that this result stemmed from a lack of agreement among this class's experts. It can be expected that the higher the agreement between the labelers, the higher the MCC of the model is attained. Consequently, it was decided to conduct a statistical agreement analysis on the ground truth labels.

\subsection{Statistical agreement analysis}

The statistical agreement analysis aims to evaluate the consistency among the ground truth labelers, who, here, are the experts. Fleiss' Kappa is a statistical measure used to assess the reliability of agreement between multiple raters for categorical items \cite{fleiss1971measuring}. Fleiss' Kappa was employed to evaluate the consistency of diagnostic decisions made by multiple experts on dental conditions

Table \ref{tab:freq-kappa-fless} contains the frequency of positive samples for each condition, the attained MCC values of the models, and the computed Fleiss' kappa values in the Expert Image Dataset. An inspection of the table's data shows that the hypothesis made holds true: the lack of agreement between the experts on condition 9 (Kappa of 0.045) was connected to the poor performance of all groups (model, students, and experts) on the same condition (MCC of 0 (zero)), indicating that the models struggled to learn from inconsistent labels. Table \ref{tab:freq-kappa-fless} also indicates substantial agreement among the labelers for conditions 1 and 6 (Kappas of 0.776 and 0.750), which are the conditions where the models attained their best results. These results suggest a correlation between the models' performance and the level of agreement among the labelers.

\begin{table*}[t]
\centering
\caption{Frequency of positive samples, Fleiss' Kappa and model's MCC on the definite evaluation set for each condition dental condition.}
\label{tab:freq-kappa-fless}
{\fontsize{2.5}{4}\selectfont
\begin{tabular}{|c|c|c|c|c|c|c|c|c|c|c|c|c|c|}
\hline
Condition & 1 & 2 & 3 & 4 & 5 & 6 & 7 & 8 & 9 & 10 & 11 & 12 & 13 \\ \hline
Frequency & 24 & 10 & 4 & 7 & 6 & 10 & 5 & 5 & 6 & 5 & 6 & 4 & 4 \\ \hline
Kappa & 0.776 & 0.506 & 0.601 & 0.479 & 0.234 & 0.750 & 0.700 & 0.256 & 0.045 & 0.389 & 0.485 & 0.336 & 0.468 \\ \hline
MCC & 0.819 & 0.749 & 0.526 & 0.577 & 0.620 & 0.755 & 0.688 & 0.264 & 0.000 & 0.369 & 0.514 & 0.397 & 0.330 \\ \hline
\end{tabular}
}
\end{table*}

\begin{figure*}[ht!p]
    \centering
    \begin{subfigure}{0.49\linewidth}
        \centering
    \includegraphics[width=\linewidth]{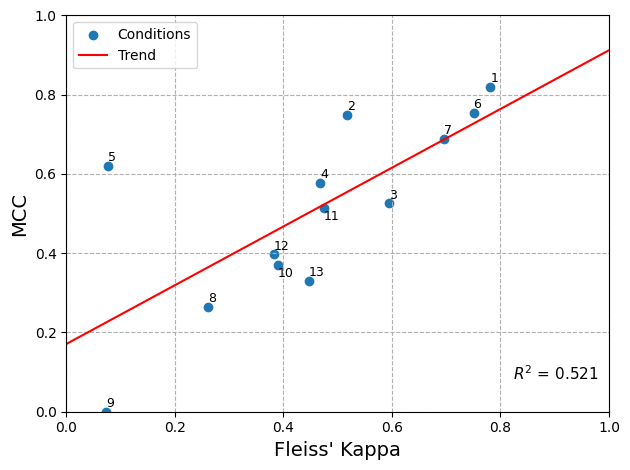}
    \caption{Scatter plot of MCC results for each condition against Fleiss' Kappa, showing an increasing trend.}
    \label{fig:kappa-mcc}
    \end{subfigure}
    \vspace{0.5cm}]
    \begin{subfigure}{0.49\linewidth}
        \centering
        \includegraphics[width=\linewidth]{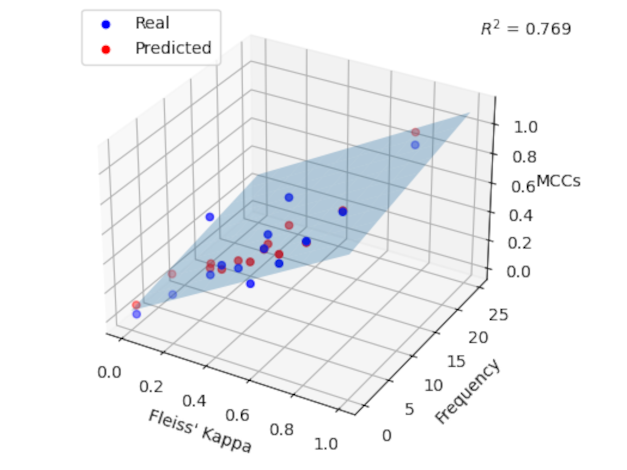}
    \caption{Scatter plot of the MCC results for each condition against Fleiss' Kappa and the frequency of positive samples in the dataset. (The blue dots represent the actual values, and the red dots represent the predicted values according to the fitted linear function.)}
    \label{fig:kappa-freq-mcc}
    \end{subfigure}
    \caption{Plots showing the linear trends of MCC results based on Fleiss' Kappa and the frequency of positive samples for each condition.}
    \label{fig:kappas}
\end{figure*}

Fig. \ref{fig:kappas}(a) is a scatter plot of the attained MCC results against the computed kappas. The value of $R^2$ reached 0.521. Similar to the correlation between the Frequency of Positive Samples and MCC of Fig. \ref{fig:mcc-trend}, this correlation was not expected to be flawless. Instead, it was aimed to demonstrate a trend—specifically, an increasing one—where a higher kappa corresponds to a higher MCC of the proposed solution.

The correlation between the independent variables (kappa values and positive sample frequency) and the dependent variable (attained MCC) was further investigated.
Fig. \ref{fig:kappas} illustrates the final result. With these two independent variables, $R^2$ reaches 0.769, a value considered a good fit.

In summary, the performed statistical agreement analysis supports the hypothesis that higher inter-rater agreement leads to better model performance, as measured by MCC. The number of positive samples also has an increasing impact on the values of MCC results. The results underscore the importance of achieving consensus among labelers to improve the reliability of ground truth data and, consequently, the performance of predictive models.

\section{Concluding Remarks}

Studies on panoramic radiographs have primarily relied on supervised learning, but this approach is becoming impractical due to its heavy dependence on annotated data. This limitation underscores the necessity to explore other learning paradigms, such as semi-supervised and self-supervised learning.

The importance of datasets in machine learning cannot be overstated, particularly in medical imaging, where classes are often highly imbalanced. Although the current study utilized the largest dataset in the literature, it still faced limitations due to size. For example, one condition had only 181 samples out of over 200,000 images, representing just 0.11\% of the total cropped image teeth. This indicates a need for even larger datasets to achieve better generalizability. Additionally, the results showed that crops containing more context yielded better outcomes, even though these sizes were chosen empirically. Future research should aim to systematically determine crop sizes and increase the number of positive samples to enhance performance.

The proposed solution's lower performance in certain conditions led to an investigation of the impact of inter-rater reliability. It was discovered that 52.1\% of model performance, as measured by MCC, correlated linearly with Fleiss' kappa. This relationship highlights the critical role of expert consensus, as higher kappa values were associated with higher MCC values. Furthermore, combining kappa with the frequency of positive examples ($R^2 = 0.769$) suggests that more extensive and more consistently labeled datasets could significantly boost performance.

In conclusion, this research advances dental image classification by addressing challenges related to imbalanced datasets. The findings emphasize the need for comprehensive datasets and consistent annotations to improve model accuracy. Moreover, exploring alternative learning paradigms can help overcome the limitations of supervised learning, paving the way for more robust and reliable dental diagnostics.

\bibliography{mybibfile}

\end{document}